\documentclass[11pt]{article}

\usepackage[preprint]{acl}

 \usepackage{microtype}

\usepackage[english,bidi=default]{babel} 
\babelfont{rm}{TeXGyreTermesX} 
\newfontfamily\hindifont{FreeSerif.otf}[Script=Devanagari]
\newfontfamily\arabicfont{Amiri-Regular.ttf}[Script=Arabic]
\newfontfamily\russianfont{DejaVuSerif.ttf}

\usepackage{amsmath}
\usepackage{amssymb}
\usepackage{booktabs} 
\usepackage{fontspec}
\usepackage{xeCJK}
\usepackage{graphicx}
\usepackage{tabularx}

\usepackage{makecell}
\usepackage{threeparttable}
\usepackage{enumitem}
\usepackage{fontawesome5}
\usepackage{subcaption}

\usepackage[table]{xcolor}
\definecolor{headergray}{RGB}{235, 240, 245}
\definecolor{rowgray}{RGB}{248, 249, 250}
\definecolor{sectiongray}{RGB}{220, 225, 230}
\definecolor{settinggray}{RGB}{110, 110, 110} 

\definecolor{affcolor}{HTML}{007575} 
\definecolor{negcolor}{HTML}{D62728} 
\definecolor{colcolor}{HTML}{7A42A5} 
\definecolor{degcolor}{HTML}{E56B00} 

\definecolor{tagbg}{HTML}{F0F0F0}
\definecolor{tagtxt}{HTML}{555555}
\definecolor{intbg}{HTML}{FCE8E8}
\definecolor{inttxt}{HTML}{D62728}

\newcommand{\tagbox}[1]{\colorbox{tagbg}{\textcolor{tagtxt}{\sffamily\fontsize{6.5}{8}\selectfont\bfseries \vphantom{Ig}#1}}}
\newcommand{\intbox}{\colorbox{intbg}{\textcolor{inttxt}{\sffamily\fontsize{6.5}{8}\selectfont\bfseries \vphantom{Ig}INT}}}
\newcommand{\gloss}[1]{\textcolor{gray}{\scriptsize\textit{#1}}}

\usepackage{tcolorbox}

\title{``Not'' is Not Lost in Translation: A Causal Analysis of Negation Representations in Multilingual LLMs}

\author{Giang Son Nguyen$^{1,\dagger}$ 
    \qquad \textbf{Wray Buntine}$^{2,3}$ 
    \qquad \textbf{Dung D. Le}$^{2}$ 
    \qquad \textbf{Wenya Wang}$^{1}$
    \\
     \textsuperscript{1} Nanyang Technological University
     \qquad
     \textsuperscript{2} VinUniversity
     \quad
     \textsuperscript{3} Monash University
     \\
        \texttt{\{ng0013on, wangwy\}@ntu.edu.sg} 
        \qquad \texttt{\{wray.b, dung.ld\}@vinuni.edu.vn}
}

\begin{document}

\maketitle
\begin{abstract}
Negation occurs in every language, yet its surface realization differs drastically.
Despite this, multilingual large language models (LLMs) translate negation robustly, raising the question of whether they share an underlying representation of negation across languages.
We study this question in six open models across English, French, Vietnamese, and Chinese, using causal interventions on a synthetic contrastive dataset and a negation translation dataset derived from FLORES-200.
Across models, we find strong evidence that negation representations are causally controllable and largely shared across languages.
In particular, we show that a contrastive activation direction extracted from one language can induce or remove negation in the others, and directions extracted from different languages are substantially aligned.
We further find that negation is represented in sparse features: a single sparse autoencoder (SAE) feature per model suffices to control polarity across languages, while responding selectively to negated inputs. 
Finally, both dense and sparse interventions generalize to four additional languages and FLORES-derived sentences.
We release data and code to support future work: \href{https://anonymous.4open.science/r/negation-translation/}{\faGithub \space negation-translation}.
\end{abstract}

\renewcommand{\thefootnote}{\fnsymbol{footnote}} 
\footnotetext{$\dagger$ Work partially done while at VinUniversity.}
\renewcommand{\thefootnote}{\arabic{footnote}}

\section{Introduction}

\begin{figure}[!t]
    \centering
    \includegraphics[
        width=\columnwidth
    ]{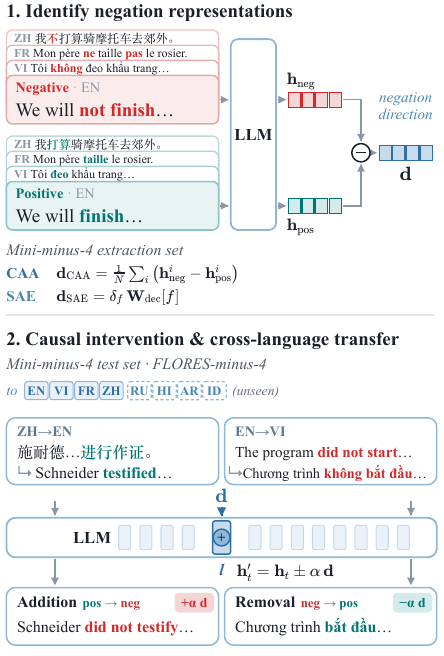}
    \caption{
    \textbf{Overview of causal interventions on negation representations}.
    Adding or subtracting a contrastive direction or one SAE feature's decoder vector during translation can reverse polarity across languages.
    }
    \label{fig:intervention_examples}
\end{figure}

\textbf{Negation} -- the reversal of a proposition's truth value, typically by inserting a word like \textit{not} -- is a universal phenomenon, yet its grammatical realization varies drastically across languages. 
Where English uses a post-copular \textit{not}, French splits it into the discontinuous \textit{ne\dots pas},
Vietnamese places the particle \textit{không} before the verb, and Chinese relies on \textit{不/没}. 
Despite these structural differences, large language models (LLMs) translate negation robustly across languages, maintaining near-perfect polarity preservation in our experiments (Appendix~\ref{app:preservation}).
This cross-lingual fluency raises a fundamental question: \emph{how do LLMs represent negation in a multilingual setting?} 

\textbf{Mechanistic interpretability} \cite{olah2020zoom, elhage2021mathematical} offers tools to study internal representations, with recent work finding evidence of shared grammatical features and refusal directions across languages \citep{brinkmann-etal-2025-large, wang2025refusal}.
However, whether negation representations are shared during translation remains less clear. \citet{brinkmann-etal-2025-large} identify features associated with negation in two LLMs, but report weak cross-lingual sharing and do not test their causal role. 
Meanwhile, \citet{zhou2026how} study negation mechanisms in monolingual English, and \citet{lasnier2026translation} analyze translation-related mechanisms without targeting specific semantic phenomena.
We instead investigate this question via \textbf{representation-based causal interventions} and \textbf{cross-lingual intervention transfer}, testing whether negation representations extracted from one language can control negation polarity in others.


To this end, we construct two complementary datasets. First, we create \textbf{Mini-minus-4}, a controlled set of minimally contrastive sentence pairs in four languages, to extract and evaluate negation representations. Second, we derive \textbf{FLORES-minus-4} from FLORES-200 \cite{nllb2024scaling} by introducing negation into naturalistic sentences, to test whether the intervention effects persist on longer, out-of-domain sentences. We release both datasets to facilitate future research.

We first identify candidate negation representations as contrastive activation directions and validate their causal role through interventions that add or remove negation during translation, using Contrastive Activation Addition (CAA)
\cite{rimsky-etal-2024-steering} (\S\ref{sec:linear}; examples in Figure~\ref{fig:intervention_examples}).
We then examine whether these directions are shared across languages through cross-language transfer and similarity analysis (\S\ref{sec:crosslingual}).
Next, we investigate whether individual sparse autoencoder (SAE) features \cite{huben2024sparse, templeton2024scaling} also capture negation and exhibit causal effects across languages (\S\ref{sec:sae}).
Finally, we test whether the intervention effects of both dense directions and sparse features generalize to unseen languages and naturalistic sentences (\S\ref{sec:generalize}).

We study six open models from three families ---
Gemma 3, Llama 3.1, and Qwen3/3.5
\cite{gemmateam2025gemma3technicalreport, llama3.1,
yang2025qwen3technicalreport, qwen3.5} ---
with English, French, Vietnamese, and Chinese as the primary languages. 
Our findings suggest that negation representations are \textbf{causally controllable, transfer strongly across languages, and exhibit substantial cross-lingual alignment}. Specifically, we show that:

\begin{itemize}
    \item \textbf{Negation representations are controllable and shared across languages}: intervening along a single contrastive direction adds or removes negation, and directions extracted from one language retain strong control in others. Per-language directions are substantially aligned.
    
    \item \textbf{Negation is represented in sparse features:} a single SAE feature can respond selectively to negated contexts and exert strong causal control over polarity across languages.
    
    \item \textbf{These causal effects generalize:} both dense directions and sparse features flip polarity in unseen languages and naturalistic sentences.
\end{itemize}

\section{Method}
\subsection{Preliminaries}

\paragraph{Decoder-only Transformers.}
We study decoder-only Transformers \citep{NIPS2017_3f5ee243}. We denote the residual stream at the output of block $l$ by
$\mathbf{h}^{l} \in \mathbb{R}^{d_{\mathrm{model}}}$, where
$d_{\mathrm{model}}$ is the model's hidden dimension.
We extract activations and apply interventions at this location.

\paragraph{Sparse Autoencoders (SAEs).}
A sparse autoencoder (SAE; \citealp{huben2024sparse,templeton2024scaling})
decomposes a residual activation
$\mathbf{h} \in \mathbb{R}^{d_{\mathrm{model}}}$ into a sparse code
$\mathbf{z} \in \mathbb{R}^{d_{\mathrm{SAE}}}$ and reconstructs it as
\begin{equation}
    \hat{\mathbf{h}}
    =
    \sum_{f=1}^{F}
    z_f\,\mathbf{W}_{\mathrm{dec}}[f]
    + \mathbf{b}_{\mathrm{dec}},
\end{equation}
where $F=d_{\mathrm{SAE}}$ is the number of SAE features, $z_f$ is the activation of feature $f$, and $\mathbf{W}_{\mathrm{dec}}[f] \in \mathbb{R}^{d_{\mathrm{model}}}$ is its decoder vector. 
Sparsity is enforced during training so that 
only a small number of features are active for any given input. 
Each decoder vector $\mathbf{W}_{\text{dec}}[f]$ can be interpreted as a direction in residual stream space 
associated with feature $f$. We intervene using these decoder vectors in \S\ref{sec:sae_method}.

\subsection{Dataset Construction}

We construct two complementary datasets. \textbf{Mini-minus-4} provides controlled minimal pairs for extracting and evaluating negation representations, while \textbf{FLORES-minus-4} tests whether their intervention effects can generalize to naturalistic sentences. Full construction steps are in Appendix~\ref{app:data-construction}.

\paragraph{Mini-minus-4} is a synthetic minimally contrastive pair dataset in English (EN), Vietnamese (VI), French (FR), and Chinese (ZH). 
Using GPT-5.4-mini \cite{gpt5.4}, we generate single-clause sentences with explicit verbal negation using the primary marker of each language: \textbf{not} (EN), \textbf{không} (VI), \textbf{ne\ldots pas} (FR), and \textbf{不} (ZH). 
Each negative sentence is paired with a positive counterpart identical in lexical content and structure, differing only in polarity. 
The purpose of this minimally contrastive setup is to isolate the negation signal using the mean difference between positive and negative activations. 
Examples of Mini-minus-4 pairs are shown in Figure~\ref{fig:extraction_example}.

We construct 350 pairs per language, split into \textbf{150 extraction / 100 evaluation / 100 test} pairs. The extraction split is used to estimate representation and steering directions, the evaluation split for hyperparameter selection and analysis, and the test split for held-out evaluation. Predicates do not overlap across splits.
To test generalization beyond the four languages used for representation extraction and selection, we additionally construct 100 test pairs each in Russian (RU), Hindi (HI), Arabic (AR), and Indonesian (ID).

\paragraph{FLORES-minus-4} is a multilingual negation translation dataset, derived from the ~{FLORES-200} devtest split \cite{nllb2024scaling} by inducing negation into the original sentences. 
We first use GPT-5.6-Luna \cite{gpt5.6} to select sentences with a main clause whose predicate can be meaningfully negated, then generate a polarity-flipped counterpart in a separate prompt. 
Finally, GPT-5.6-Terra verifies that the result is grammatical and preserves the original meaning apart from polarity.
Since the dataset supports any-to-any translation, we retain an item only when it is verified in \textit{all four languages}, yielding \textbf{216 items}.

\paragraph{Activation extraction.}
At layer $l$, we extract residual-stream activations at the output of decoder block $l$.
We consider two token positions (Figure~\ref{fig:extraction_example}): the \textbf{negated predicate} (NP), the first token of the negated verb and its corresponding positive counterpart\footnote{In Chinese, where the negator merges with the verb into a single token (e.g.\ 不高), the merged token is used.}, and the \textbf{final source token} (FST), typically the period.

\newcommand{\NPneg}[1]{\textcolor{negcolor}{\textbf{\underline{#1}}}}
\newcommand{\NPpos}[1]{\textcolor{affcolor}{\textbf{\underline{#1}}}}

\begin{figure}[t]
\centering
\footnotesize
\setlength{\tabcolsep}{2.0pt}
\renewcommand{\arraystretch}{1.2}

\begin{tabularx}{\linewidth}{@{}l
    >{\raggedright\arraybackslash}X
    >{\raggedright\arraybackslash}X@{}}
\toprule
\textbf{Lang.} & \textbf{Negative} & \textbf{Positive} \\
\midrule

EN &
We will \textcolor{negcolor}{not}
\NPneg{finish} the chapter tonight. &
We will \NPpos{finish} the chapter tonight. \\

VI &
Tôi \textcolor{negcolor}{không}
\NPneg{đeo} khẩu trang khi ra ngoài. &
Tôi \NPpos{đeo} khẩu trang khi ra ngoài. \\

FR &
Mon père \textcolor{negcolor}{ne}
\NPneg{taille} \textcolor{negcolor}{pas} le rosier. &
Mon père \NPpos{taille} le rosier. \\

ZH &
我\textcolor{negcolor}{不}\NPneg{打算}骑摩托车去郊外。 &
我\NPpos{打算}骑摩托车去郊外。 \\

\bottomrule
\end{tabularx}

\caption{\textbf{Examples of Mini-minus-4 contrastive pairs
and activation extraction positions.}
Each pair differs only in polarity.
The negated predicate (NP) position is bold and underlined.
The final source token (FST) is typically the period.}
\label{fig:extraction_example}
\end{figure}

\subsection{Causal Intervention with Contrastive Directions}
\label{sec:caa_method}

We first identify candidate negation representations and validate their causal role by intervening on them during generation (i.e. using activation steering). We employ \textbf{Contrastive Activation Addition} (CAA; \citealp{rimsky-etal-2024-steering}). 
For each model, layer $l$, position $p \in \{\mathrm{NP}, \mathrm{FST}\}$, and language $L$, we compute a negation direction from the $N_{\mathrm{extract}}=150$ Mini-minus-4 extraction pairs:
\[
\mathbf{d}_{\mathrm{CAA}}^{l,p}
=
\frac{1}{N}
\sum_{i=1}^{N}
\left(
\mathbf{h}^{l,p,i}_{\mathrm{neg}}
-
\mathbf{h}^{l,p,i}_{\mathrm{pos}}
\right),
\]
where $\mathbf{h}^{l,p,i}_{\mathrm{neg}}$ and $\mathbf{h}^{l,p,i}_{\mathrm{pos}}$ are the activations of the negative and positive members of pair $i$. We omit the language index $L$ for simplicity unless explicitly needed.
We refer to the directions extracted across layers, positions, and languages collectively as \textit{negation directions}.

During generation, we intervene at a single layer by adding this direction to the activation
\[
\mathbf{h}'_t
=
\mathbf{h}_t
+
\alpha_{\mathrm{CAA}}
\mathbf{d}_{\mathrm{CAA}}.
\]
Because $\mathbf{d}_{\mathrm{CAA}}$ points from positive to negative polarity, $\alpha>0$ adds negation and $\alpha<0$ removes it; we refer to either operation as a \textbf{flip}. We sweep $|\alpha_{\mathrm{CAA}}| \in \{2,4,6\}$.

The intervention is applied to all source-sentence tokens, excluding the instruction and generated output. On FLORES-minus-4, we annotate positions of the main clause's tokens and intervene only on those positions.


\subsection{Causal Intervention with Sparse Features}
\label{sec:sae_method}

We next test whether negation is also represented in SAE features. For each feature $f$, we compute its mean activation difference between negative and positive examples at the NP position:\footnote{We use NP because it gives more consistent steering results in preliminary experiments.}
\[
\delta_f
=
\mu_{\mathrm{neg},f}
-
\mu_{\mathrm{pos},f}.
\]
We rank features by $\delta_f$ and retain the top-$K$ features $\mathcal{F}_K$, thereby selecting features that activate more strongly on negative than positive examples.

We construct an SAE steering direction from their decoder vectors:
\[
\mathbf{d}_{\mathrm{SAE}}^{l,p,K}
=
\sum_{f\in\mathcal{F}_K}
\delta_f\,\mathbf{W}_{\mathrm{dec}}[f].
\]

We intervene using the same sign convention and scope as CAA:
\[
\mathbf{h}'_t
=
\mathbf{h}_t
+
\alpha_{\mathrm{SAE}}
\mathbf{d}_{\mathrm{SAE}}.
\]

\paragraph{Norm matching.}
Because CAA and SAE directions differ substantially in norm, their steering coefficients are not directly comparable. For an SAE direction at layer $l$ and position $p$, we use the best CAA coefficient selected on EN at the same $(l,p)$ (Appendix~\ref{app:norm-matching}), denoted $\alpha_{\mathrm{CAA}}^{*}(l,p)$, and set
\[
|\alpha_{\mathrm{SAE}}|
=
|\alpha_{\mathrm{CAA}}^{*}(l,p)|
\frac{
\left\|\mathbf{d}_{\mathrm{CAA}}^{l,p,\mathrm{EN}}\right\|
}{
\left\|\mathbf{d}_{\mathrm{SAE}}^{l,p,\mathrm{EN},K}\right\|
}.
\]

\begin{figure*}
    \centering
    \includegraphics[width=1\linewidth]{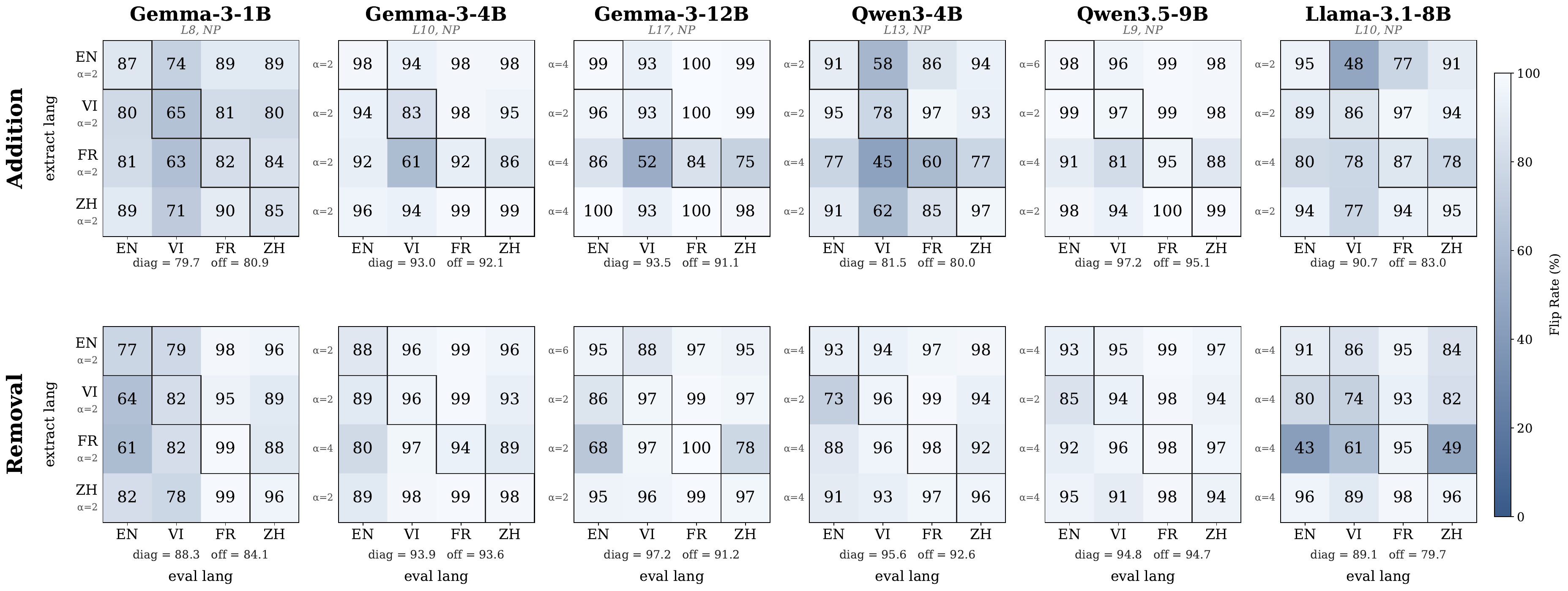}
    \caption{\textbf{Cross-language transfer flip rates \textnormal{(off-diagonal cells)} for CAA steering} (top: addition, bottom: removal). 
    Flip rates remain moderate to high when applying negation directions from any language to the others, indicating \textit{negation directions transfer strongly across languages}.
    }
    \label{fig:dir_crosslang_transfer}
\end{figure*}

\subsection{Intervention Success Evaluation}

We evaluate intervened translations using four criteria. 
Thresholds and judge prompts are provided in Appendix~\ref{app:steering-eval}.

\begin{enumerate}
\item \textbf{Polarity.} The output exhibits the intended polarity change. For Mini-minus-4, we use language-specific negation-marker detection rules. For FLORES-minus-4, GPT-5.6-Terra judges whether the intervened output is negated or affirmative.

\item \textbf{Target language.} fastText language identification \cite{joulin2016fasttextzipcompressingtextclassification,joulin-etal-2017-bag} verifies that the output is in the requested target language.

\item \textbf{Non-degeneracy.} The output satisfies minimum token-count and unique-token-ratio thresholds, filtering out degenerate or repetitive generations.

\item \textbf{Content preservation.} For Mini-minus-4, the output must exceed a chrF threshold \cite{popovic-2015-chrf} against the model's baseline translation of the opposite-polarity counterpart.\footnote{We compare against the model's own baseline output so that this criterion measures wording changes introduced by steering rather than divergence from a human reference.}
For FLORES-minus-4, GPT-5.6-Terra judges whether the baseline and steered translations describe the same underlying content except for the polarity difference.
\end{enumerate}

We compute the \textbf{Flip Rate} as the percentage of eligible examples that satisfy all four criteria. An example is eligible only if the un-intervened baseline translation preserves the source polarity. This prevents attributing incorrect baseline polarity to the intervention. Where necessary, we refer to this metric as the \textbf{Addition Rate} (positive $\xrightarrow[]{}$ negative) and \textbf{Removal Rate} (negative $\xrightarrow[]{}$ positive).

\paragraph{Human verification.} 
We verify our automatic evaluation with human evaluators (details in Appendix~\ref{app:human_validation_judge}). 
Evaluators independently judged all steered outputs of Gemma-3-4B on both datasets, covering all languages and both flip modes.
Over four languages, human evaluations agree with the rule-based evaluator on Mini-minus-4 in 96.5\% and with the LLM judge on FLORES-minus-4 in 93.7\% of samples.
These high agreement rates support the reliability of the automated \textbf{Flip Rates}, which we use throughout our experiments.

\section{Setup}

\subsection{Models}
We evaluate 6 instruction-tuned models spanning 3 families: Gemma~3~(1B, 4B, 12B)~IT, Qwen3~4B, Qwen3.5~9B, and Llama~3.1~8B~Instruct \cite{gemmateam2025gemma3technicalreport, yang2025qwen3technicalreport, qwen3.5, llama3.1}. 
We drop the -IT and -Instruct suffixes when referring to the models in text.

For each family, we choose one model for which pretrained SAEs are publicly available to perform SAE-based steering: Gemma-3-4B \cite{mcdougall2025gemmascope2}, Llama-3.1-8B \cite{he2024llamascopeextractingmillions}, and Qwen3.5-9B \cite{deng2026qwenscopeturningsparsefeatures}. Details of SAEs used in our experiments are in Appendix~\ref{app:sae-specs}.

\subsection{Translation task}

All experiments use a zero-shot translation prompt:
\begin{tcolorbox}[colback=gray!5!white, colframe=gray!60!black, boxrule=0.5mm, arc=2mm]
\small
  \textsf{Translate the following sentence into \textbf{\{TargetLanguage\}}. Only output the translation.}\\[1ex]
  \textbf{Sentence:} \texttt{\{sentence\}}
\end{tcolorbox}

For the main direction extraction and intervention experiments, we consider four translation settings: EN$\xrightarrow{}$VI, VI$\xrightarrow{}$EN, FR$\xrightarrow{}$EN, ZH$\xrightarrow{}$EN. 
For the additional-language generalization test in \S\ref{sec:generalize}, we evaluate \{RU,HI,AR,ID\}$\xrightarrow{}$EN.

We find that varying the target language produces nearly identical negation directions and similar intervention results (Appendix~\ref{app:varying_target}), therefore we only vary the source language in our main experiments.
We refer to translation settings by the source languages for brevity.

\section{Results}

\newcommand{\figstub}[1]{%
  \fbox{\parbox[c][3.4cm][c]{0.97\linewidth}{\centering\ttfamily\footnotesize #1}}}
\newcommand{\setting}[1]{{\color{settinggray}\footnotesize #1}}



\subsection{Negation directions causally control polarity}
\label{sec:linear}

For each model, we sweep layer-position pairs $(l,p)$ on the EN evaluation set with $l$ ranging from 10-55\% of model depth. We select a single $(l,p)$ maximizing addition/removal performance; this pair is then fixed across languages. 
At the selected $(l,p)$, we compute a direction $\mathbf{d}^{l,p}$ for each language $L$ and separately select $|\alpha| \in \{2,4,6\}$ for addition and removal on that language's evaluation set. 
Intervention results of selected $(l,p)$ and $\alpha$ on the test set are reported in Table~\ref{tab:direction_peaks}. 
Full sweeps and selected values are reported in Appendix~\ref{app:caa_flip}.

\begin{table}[h]
\centering
\fontsize{8pt}{9pt}\selectfont
\setlength{\tabcolsep}{2.3pt}

\renewcommand{\arraystretch}{1.3}

\begin{tabular}{l cccc}
\toprule
\rowcolor{headergray}
\textbf{Model} & \textbf{EN} & \textbf{VI} & \textbf{FR} & \textbf{ZH} \\
\midrule

Gemma-3-1B
& \textbf{86.9} / 77.0
& 65.0 / \textbf{81.8}
& \textbf{82.0} / \textbf{99.0}
& \textbf{85.0} / \textbf{95.5} \\

\rowcolor{rowgray}
Gemma-3-4B
& \textbf{98.0} / \textbf{88.0}
& \textbf{83.0} / \textbf{95.9}
& \textbf{92.0} / \textbf{94.0}
& \textbf{99.0} / \textbf{97.9} \\

Gemma-3-12B
& \textbf{99.0} / \textbf{95.0}
& \textbf{93.0} / \textbf{97.0}
& \textbf{84.0} / \textbf{100.0}
& \textbf{98.0} / \textbf{96.9} \\

\rowcolor{rowgray}
Qwen3-4B
& \textbf{90.9} / \textbf{92.7}
& 78.0 / \textbf{95.9}
& 60.0 / \textbf{98.0}
& \textbf{97.0} / \textbf{95.7} \\

Qwen3.5-9B
& \textbf{98.0} / \textbf{93.0}
& \textbf{97.0} / \textbf{93.9}
& \textbf{95.0} / \textbf{98.0}
& \textbf{99.0} / \textbf{94.5} \\

\rowcolor{rowgray}
Llama-3.1-8B
& \textbf{94.9} / \textbf{91.0}
& \textbf{86.0} / 74.5
& \textbf{87.0} / \textbf{95.0}
& \textbf{95.0} / \textbf{95.8} \\

\bottomrule
\end{tabular}

\caption{\textbf{Within-language CAA addition/removal rates.}
($l,p$) is chosen once, while $\alpha$ is chosen for each language and flip mode by value sweeps over the evaluation set.
Values $\geq80\%$ are bolded.
}
\label{tab:direction_peaks}
\end{table}

\textbf{Adding or subtracting ${\mathbf{d}_\text{CAA}}$ reverses polarity in the translation with high success}, providing causal evidence for a negation representation in the model.
Negation directions on Gemma-3-(4B, 12B) and Qwen3.5-9B achieve $\geq80\%$ flip rates on all languages.
Addition rates on Qwen3-4B drop to 60.0\% on FR, while Gemma-3-1B's VI addition is 65.0\%.
Llama-3.1-8B is strong for addition across languages, but its VI removal rate drops to 74.5\%.
These flip rates are far above those of random directions (typically 0\%, Appendix~\ref{app:random_control}).

\textbf{The effect also extends beyond explicit negation markers.}
During manual inspection, we found that 44 ZH negative examples across six models were translated using lexical antonyms or negative prefixes rather than explicit negation (\textcolor{negcolor}{\textit{不愿意}}$\xrightarrow[]{}$\textcolor{negcolor}{\textit{unwilling}}, Figure~\ref{fig:morphological_negation}). Subtracting the negation direction reverses the meaning in 68.2\% of those cases (\textcolor{negcolor}{\textit{unwilling}} $\xrightarrow[]{}$ \textcolor{affcolor}{\textit{want}}, Appendix~\ref{app:eligible_count}).

\begin{figure*}
    \centering
    \includegraphics[width=1\linewidth]{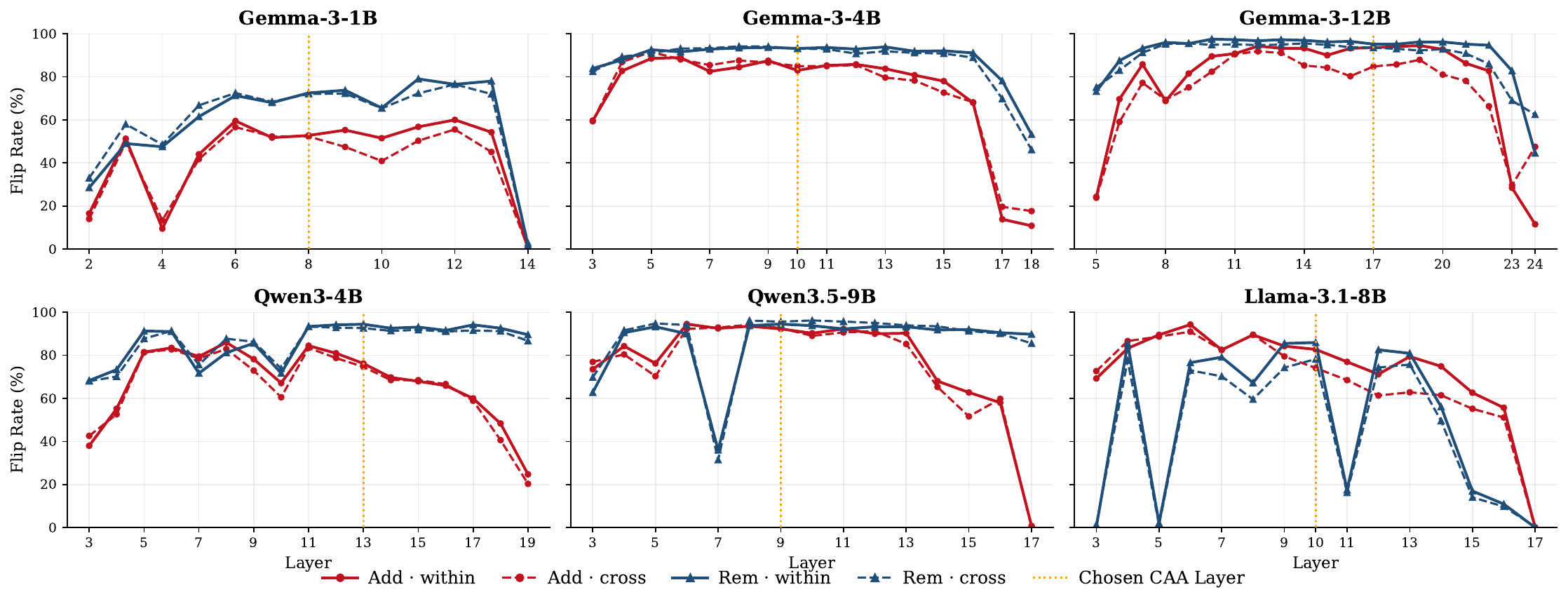}
    \caption{\textbf{Per-layer comparison of within and cross-language flip rates.} Cross-language lines are means of all 12 transfer directions' flip rates, within-language lines are means of 4 within-language flip rates. \textit{Across all models, cross-language transfer strength follows within-language control and remains strong in multiple layers}.
    }
    \label{fig:crosslang_within_vs_cross_grid}
\end{figure*}

\begin{figure}[t]
\centering
\small
\renewcommand{\arraystretch}{1.05}
\begin{tabularx}{\linewidth}{@{} >{\raggedright\arraybackslash}X @{}}
\toprule
\textbf{Negation Removal beyond negation markers} (ZH$\to$EN) \\
\midrule
\tagbox{SRC} 我们\textcolor{negcolor}{\textbf{不愿意}}参加讨论。 \gloss{(unwilling)} \\
We are \textcolor{negcolor}{\textbf{unwilling}} to participate\ldots\ $\to$ We \textcolor{affcolor}{\textbf{want}} to participate\ldots \\
\midrule
\tagbox{SRC} 同事们\textcolor{negcolor}{\textbf{不常}}点外卖。 \gloss{(rarely)} \\
Colleagues \textcolor{negcolor}{\textbf{rarely}} order takeout. $\to$ Colleagues \textcolor{affcolor}{\textbf{often}} order takeout. \\
\bottomrule
\end{tabularx}
\caption{\textbf{Intervention flips negation without explicit markers} (Qwen3-4B). Baseline $\to$ steered translation.}
\label{fig:morphological_negation}
\end{figure}

\subsection{Negation directions align and transfer across languages}
\label{sec:crosslingual}

To test whether negation directions are shared across languages, we apply the direction and $\alpha$ selected for one language $L$ to test examples in another language $L'$ at the same $(l,p)$. Figure~\ref{fig:dir_crosslang_transfer} reports the resulting test-set transfer matrices, with within-language flip rates on the diagonal and cross-language transfer off the diagonal.

Across all models, \textbf{a single EN-extracted direction is enough to flip polarity in all four languages}. 
In particular, $\mathbf{d}^{\mathrm{EN}}$ achieves 48--100\% addition and 77--99\% removal rates across languages. Transfer is particularly strong for Gemma-3-4B, Gemma-3-12B, and Qwen3.5-9B, where $\mathbf{d}^{\mathrm{EN}}$ achieves 93--100\% addition and 88--99\% removal on non-EN languages.

\textbf{More broadly, directions extracted from any language transfer to the others}. 
Across models, mean transfer rates range from 80.0--95.1\% for addition and 79.7--94.7\% for removal. 
Transfer strength varies by language pair: addition is weakest to VI, especially for $\mathbf{d}^{\text{FR}}$ (45--81\%), while Llama-3.1-8B's $\mathbf{d}^{\text{FR}}$ is also weak for removal on EN, VI, and ZH (43\%, 61\%, and 49\%).

Next, we ask whether cross-language transfer is specific to the selected layer. On the evaluation set, we sweep layers for all models while fixing $p=\mathrm{NP}$ and $\alpha=4$, and compare the mean within-language flip rate with the mean cross-language flip rate at each layer. Cross- and within-language flip rates track each other closely (Figure~\ref{fig:crosslang_within_vs_cross_grid}), suggesting that \textbf{where negation directions are causally controllable, they also transfer across languages}.
Furthermore, within and cross-language flip rates remain stably strong in early-to-mid layers of most models except Llama-3.1-8B, indicating this is not an artifact of the chosen layer.

\begin{figure}[t]
    \centering
    \includegraphics[width=\linewidth]{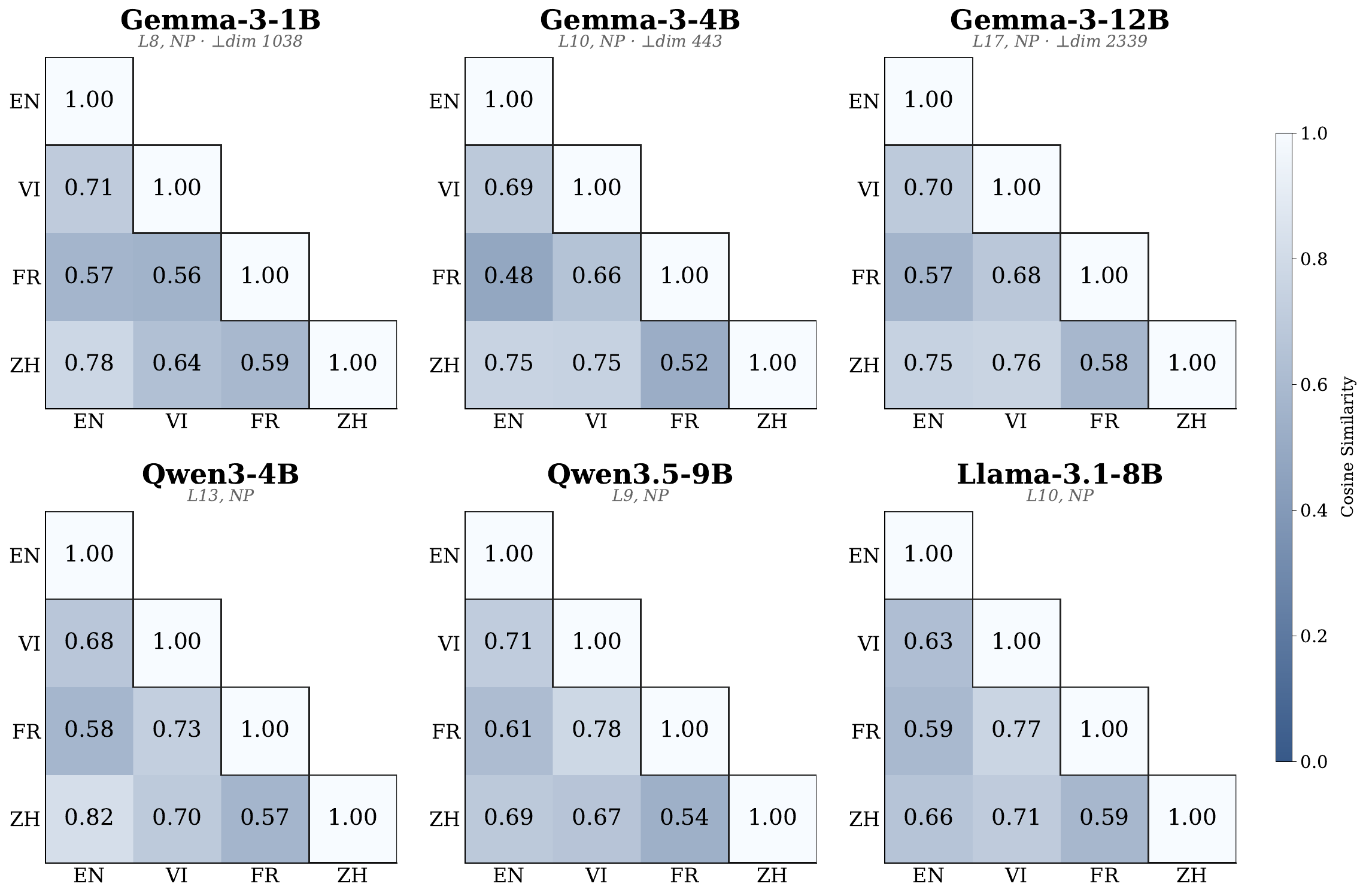}
    \caption{
    \textbf{Pairwise cosine similarities between negation directions}
    extracted from different languages at the same $(l,p)$.}
    \label{fig:crosslang_cosine_heatmap}
\end{figure}

Finally, we analyze the similarity of negation directions extracted from each language at the same $(l,p)$. For Gemma models, we zero out dominant massive-activation dimensions per model before computing similarities; we discuss how they distort similarity in Appendix~\ref{app:massive_activation}. 
In every model, \textbf{negation directions from all languages are positively aligned}, with pairwise cosine similarities ranging from 0.48 to 0.82 (Figure~\ref{fig:crosslang_cosine_heatmap}).
Together, these results provide evidence of \textbf{cross-lingual alignment in negation representations}.

\subsection{Negation is represented in sparse features}
\label{sec:sae}

\begin{figure*}
    \centering
    \includegraphics[width=0.9\linewidth]{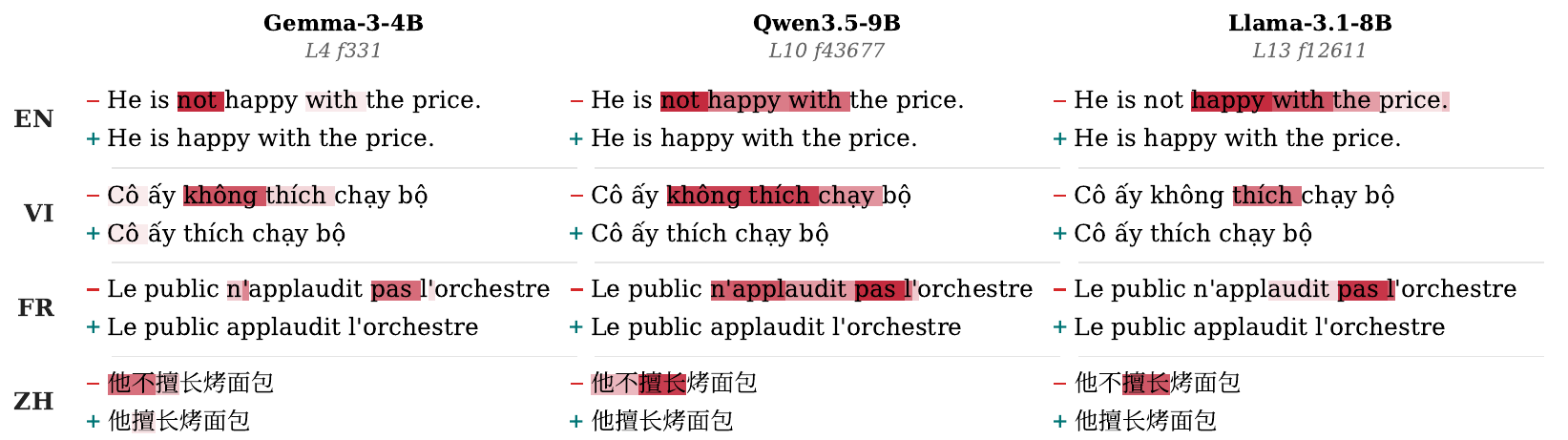}
    \caption{\textbf{Representative activation patterns of the selected SAE features.}
    Darker red indicates higher SAE activation.
    Gemma-3-4B L4 f331 fires strongly on positions of negation markers and Llama-3.1-8B L13 f12611 activates on tokens in negated contexts, while Qwen3.5-9B L10 f43677 fires on both.}
    \label{fig:grid_features_x_langs}
\end{figure*}

We next ask whether negation is also captured by individual SAE features. Because exhaustive feature evaluation is costly, we first identify candidate layers using EN.
At each layer, we steer with the top-$K$=10 features (ranked by $\delta_f$) and retain layers with strong addition/removal rates on the EN evaluation set. Within these layers, we evaluate individual features ($K$=1), select one EN feature per model, and apply it to the test sets of all four languages. 
Full selection results are in Appendix~\ref{app:sae}.

For every tested model, we find \textbf{individual SAE features whose decoder directions strongly control polarity across languages} (Table~\ref{tab:sae_k1_crosslang_transfer}). Gemma-3-4B's L4 f331\footnote{L4 f331 refers to feature 331 in the layer-4 SAE of Gemma-3-4B.} and Qwen3.5-9B's L10 f43677 achieve $\geq91\%$ and $\geq93\%$ flip rates, respectively, for both addition and removal across all four languages.
Llama-3.1-8B's L13 f12611 is weaker on VI, but still achieves strong control in the other languages.

\begin{table}[h]
\centering
\fontsize{8.5pt}{9pt}\selectfont
\setlength{\tabcolsep}{2.3pt}
\renewcommand{\arraystretch}{1.25}

\begin{tabular}{@{}l cccc@{}}
\toprule
\rowcolor{headergray}
\textbf{Feature} & \textbf{EN} & \textbf{VI} & \textbf{FR} & \textbf{ZH} \\
\midrule

G4B L4 f331
& \textbf{96.0}\,/\,\textbf{91.0}
& \textbf{91.0}\,/\,\textbf{95.9}
& \textbf{97.0}\,/\,\textbf{95.0}
& \textbf{95.0}\,/\,\textbf{94.8} \\

\rowcolor{rowgray}
Q9B L10 f43677
& \textbf{97.0}\,/\,\textbf{95.0}
& \textbf{93.0}\,/\,\textbf{93.9}
& \textbf{99.0}\,/\,\textbf{97.0}
& \textbf{96.0}\,/\,\textbf{97.8} \\

L8B L13 f12611
& \textbf{83.8}\,/\,\textbf{91.0}
& 72.0\,/\,57.1
& \textbf{87.0}\,/\,\textbf{92.0}
& \textbf{90.0}\,/\,\textbf{82.3} \\

\bottomrule
\end{tabular}

\caption{\textbf{Addition/removal rates using one top-performing
EN-selected SAE feature ($K=1$) per model.}
Values $\geq 80\%$ are bolded.}
\label{tab:sae_k1_crosslang_transfer}
\end{table}

We inspect the activation patterns to characterize the three features (examples shown in Figure~\ref{fig:grid_features_x_langs}, full details included in Appendix~\ref{app:sae_feature_interpretation}).
We find that these features respond differently to negation cues: Gemma-3-4B L4 f331 acts as a \textit{cross-lingual negation-marker detector}, activating strongly on positions of negation markers, while Llama-3.1-8B L13 f12611 \textit{activates on tokens in negated contexts}.
Meanwhile, Qwen3.5-9B L10 f43677 fires on \textit{both negation markers and negated contexts}. Notably, \textbf{these features largely separate negated inputs from affirmatives across languages}. One exception is Llama's L13 f12611 firing less frequently on VI, which may explain its weak control.


\subsection{Negation directions and features generalize to unseen languages and naturalistic sentences}
\label{sec:generalize}

Finally, we test whether the representations identified on Mini-minus-4 generalize beyond the original four languages and the controlled synthetic setting.
For each model, we fix \textbf{one EN-derived contrast direction $\mathbf{d}^{\mathrm{EN}}$}, and \textbf{one EN-selected SAE feature} from \S\ref{sec:crosslingual} and \S\ref{sec:sae}, respectively, without any re-selection.
We evaluate these interventions first on RU, HI, AR, and ID test sets, and then on all four original languages in FLORES-minus-4.

\begin{table}[h]
\centering
\fontsize{8pt}{9pt}\selectfont
\setlength{\tabcolsep}{2pt}
\renewcommand{\arraystretch}{1.22}

\begin{subtable}{\linewidth}
\centering
\begin{tabular}{@{}l cccc@{}}
\toprule
\rowcolor{headergray}
\textbf{Model / Feature} & \textbf{RU} & \textbf{HI} & \textbf{AR} & \textbf{ID} \\
\midrule

\multicolumn{5}{@{}l}{\textbf{CAA}} \\

Gemma-3-4B
& \textbf{98.0}\,/\,\textbf{99.0}
& \textbf{97.0}\,/\,\textbf{93.9}
& \textbf{100.0}\,/\,\textbf{97.0}
& \textbf{99.0}\,/\,\textbf{98.0} \\

\rowcolor{rowgray}
Qwen3.5-9B
& \textbf{99.0}\,/\,\textbf{96.0}
& \textbf{98.0}\,/\,\textbf{96.0}
& \textbf{100.0}\,/\,\textbf{95.0}
& \textbf{98.0}\,/\,\textbf{99.0} \\

Llama-3.1-8B
& \textbf{97.9}\,/\,\textbf{91.9}
& \textbf{89.0}\,/\,\textbf{88.0}
& \textbf{95.9}\,/\,\textbf{94.9}
& \textbf{85.6}\,/\,\textbf{88.5} \\

\midrule

\multicolumn{5}{@{}l}{\textbf{SAE}} \\

G4B L4 f331
& \textbf{96.0}\,/\,\textbf{99.0}
& \textbf{95.0}\,/\,\textbf{86.9}
& \textbf{97.0}\,/\,\textbf{95.0}
& \textbf{92.9}\,/\,\textbf{97.0} \\

\rowcolor{rowgray}
Q9B L10 f43677
& \textbf{96.0}\,/\,\textbf{95.0}
& \textbf{97.0}\,/\,\textbf{92.0}
& \textbf{99.0}\,/\,\textbf{96.0}
& \textbf{93.0}\,/\,\textbf{99.0} \\

L8B L13 f12611
& \textbf{91.5}\,/\,\textbf{94.9}
& \textbf{86.6}\,/\,\textbf{94.3}
& \textbf{92.8}\,/\,\textbf{92.2}
& 77.2\,/\,\textbf{91.9} \\

\bottomrule
\end{tabular}
\caption{\textbf{Additional languages.}}
\label{tab:generalization_unseen}
\end{subtable}

\vspace{4pt}

\begin{subtable}{\linewidth}
\centering
\begin{tabular}{@{}l cccc@{}}
\toprule
\rowcolor{headergray}
\textbf{Model / Feature} & \textbf{EN} & \textbf{VI} & \textbf{FR} & \textbf{ZH} \\
\midrule

\multicolumn{5}{@{}l}{\textbf{CAA}} \\

Gemma-3-4B
& \textbf{91.2}\,/\,\textbf{99.5}
& \textbf{92.5}\,/\,\textbf{99.5}
& \textbf{94.0}\,/\,\textbf{99.5}
& \textbf{95.4}\,/\,\textbf{99.0} \\

\rowcolor{rowgray}
Qwen3.5-9B
& \textbf{85.6}\,/\,\textbf{98.6}
& \textbf{88.3}\,/\,\textbf{97.6}
& \textbf{88.4}\,/\,\textbf{98.6}
& \textbf{84.2}\,/\,\textbf{98.0} \\

Llama-3.1-8B
& \textbf{85.6}\,/\,\textbf{88.4}
& 58.4\,/\,65.0
& 66.7\,/\,\textbf{94.9}
& 62.6\,/\,\textbf{86.5} \\

\midrule

\multicolumn{5}{@{}l}{\textbf{SAE}} \\

G4B L4 f331
& \textbf{95.3}\,/\,\textbf{99.5}
& \textbf{91.5}\,/\,\textbf{99.5}
& \textbf{97.2}\,/\,\textbf{99.5}
& \textbf{94.0}\,/\,\textbf{99.5} \\

\rowcolor{rowgray}
Q9B L10 f43677
& \textbf{95.8}\,/\,\textbf{99.5}
& \textbf{91.1}\,/\,\textbf{99.0}
& \textbf{95.8}\,/\,\textbf{99.5}
& \textbf{95.3}\,/\,\textbf{98.0} \\

L8B L13 f12611
& 78.2\,/\,\textbf{97.7}
& 63.3\,/\,\textbf{91.6}
& 67.9\,/\,\textbf{96.3}
& 62.9\,/\,\textbf{91.8} \\

\bottomrule
\end{tabular}
\caption{\textbf{FLORES-minus-4.}}
\label{tab:generalization_flores}
\end{subtable}

\caption{\textbf{Generalization of dense and sparse negation interventions.}
Each cell reports addition/removal rates.
CAA uses the $\mathbf{d}^\text{EN}$ for each model, while SAE uses the single EN-selected feature. Values $\geq80\%$ are bolded.}
\label{tab:generalization}
\end{table}

We find that \textbf{both dense and sparse interventions generalize to unseen languages} (Table~\ref{tab:generalization}).
Contrastive directions transfer strongly, with addition/removal rates $\ge$ 85.6/88.0\% across all models.
Single SAE features likewise maintain $\geq86.9\%$ across Gemma-3-4B and Qwen3.5-9B, while Llama-3.1-8B remains $\ge$ 77.2\%.

Importantly, \textbf{both dense negation directions and single SAE features generalize to naturalistic sentences} (Table~\ref{tab:generalization}).
Removal is robust: SAE achieves $\geq91.6\%$ across all models and languages, while CAA is $\geq86.5\%$ except for Llama-3.1-8B on VI (65.0\%).
Addition remains strong for Gemma-3-4B and Qwen3.5-9B ($\geq84.2\%$), but is weaker for Llama-3.1-8B.

\section{Related Work}
\paragraph{Negation in NLP and machine translation.}

Language models often under-utilize negation cues \citep{kassner-schutze-2020-negated, ettinger-2020-bert}, and modern LLMs continue to struggle on negation-focused benchmarks \citep{hossain-etal-2020-analysis, truong-etal-2023-language, garcia-ferrero-etal-2023-dataset}.
Dropped or inverted polarity is also a long-documented failure mode in machine translation \citep{fancellu-webber-2015-translating, tang-etal-2021-revisiting}. However, this literature is largely behavioral, leaving how multilingual LLMs internally represent negation during translation underexplored.

\paragraph{Linear representations and activation steering.}

The linear representation hypothesis posits that high-level concepts correspond to directions in activation space \citep{park2024the}. 
Causally effective directions have been identified for truth, sentiment, and refusal \citep{marks2024the, hollinsworth-etal-2024-language, arditi2024refusal}, and can be manipulated through activation steering \citep{turner2024steeringlanguagemodelsactivation, rimsky-etal-2024-steering}. In this work, we employ activation steering to validate the causal role of identified representations.

\paragraph{Sparse feature decomposition.}
Sparse autoencoders (SAEs) decompose dense activations into sparse feature dictionaries \citep{huben2024sparse, templeton2024scaling}, with pretrained SAEs released for several major model families \citep{mcdougall2025gemmascope2, he2024llamascopeextractingmillions, deng2026qwenscopeturningsparsefeatures}. 
In multilingual settings, \citet{al-ghussin-etal-2026-multilingual} show that SAEs trained on multilingual data yield more reliable cross-lingual steering,
while \citet{nguyen2026recurrenceenoughcausallyvalidating} find that most SAE features do not causally transfer across languages despite recurring activations.
Whether SAE features can causally control desired behaviors as effectively as dense directions remains contested \citep{wu2025axbench, jørgensen2026steeringllmsactuallysparse, arad-etal-2025-saes}. 

\paragraph{Cross-lingual structure in multilingual LLMs.}
Evidence for language-neutral representations in multilingual LLMs is mixed. Some work finds English-centric or dominant-language semantic spaces \citep{wendler-etal-2024-llamas, wu2025the}, while \citet{brinkmann-etal-2025-large} find shared latent grammatical concepts across typologically diverse languages using SAEs.
Causal studies similarly range from a universal refusal direction \citep{wang2025refusal} to language-dependent steering effectiveness \citep{schut2025do}.


\paragraph{Adjacent mechanistic work.}

\citet{zhou2026how} study negation mechanisms in monolingual English, while \citet{lasnier2026translation} analyze mechanisms supporting machine translation. 
The former does not study cross-lingual translation or causal polarity control, and the latter targets whether and in which language translation occurs rather than specific semantic content. 
\citet{brinkmann-etal-2025-large} identify SAE features associated with negative polarity, but find weak cross-lingual sharing. They also do not examine dense directions or test the causal role of these features.
We instead investigate whether negation representations are causally shared through cross-language interventions on both dense directions and sparse features.

\section{Conclusion}
In this work, we investigated how multilingual LLMs represent negation during translation in four main languages. 
We showed that negation can be causally controlled in multilingual LLMs through both dense activation directions and individual SAE features. 
Importantly, these representations transfer across languages despite large differences in how negation is expressed, generalizing both to naturalistic sentences and to languages unseen during representation extraction.
Similarity analysis revealed negation directions are substantially similar, and activation inspection showed SAE features consistently responding to negation in every language.
These findings suggest that multilingual LLMs encode negation in a largely shared and causally controllable representation.

\section{Limitations}
Our extraction and interventions focus on verbal negation marked by explicit lexical cues (e.g., \textit{not}, \textit{không}, \textit{ne\ldots pas}). 
We also focus only on the translation task. 
Morphological negation (e.g., \textit{un-}, \textit{im-}), implicit negative expressions (e.g., \textit{fail}, \textit{lack}) and the role of negation representations in other tasks remain for future work.


Activation addition, like other linear steering methods, depends on the $\alpha$ coefficient (the steering strength), which is a known limitation in the steering literature \cite{vu2025angular,you2026spherical,aparin2026a}.
We select $\alpha$ empirically by sweeping values on the evaluation set rather than a principled selection.

For sparse feature analysis, we rely on pre-trained SAE releases, thus our analysis does not cover all models. Our feature-level interpretation is also limited to one selected feature per model.

\section*{Acknowledgments}

This research is supported by the National Research Foundation, Singapore under its National Large Language Models Funding Initiative, (AISG Award No: AISG-NMLP-2024-005). It is also supported by the MOE AcRF Tier 1 Seed Grant (RS37/24, \#025041-00001), Singapore and the $\text{VinUniversity}$ Cross-College Research Grant (Grant ID: VUNI.2324.CC06).

Giang Son Nguyen thanks the AUMOVIO-NTU Corporate Lab for lending their GPU resources for the experiments in this paper.

\bibliography{custom}

\section*{Appendix}
\appendix

\section{Additional Analysis}

\begin{table*}[h]
    \centering
    \setlength{\tabcolsep}{4pt}
    \begin{tabular}{l ccc cc}
        \toprule
        \rowcolor{headergray}
        \textbf{Model} & \textbf{Dim. Index} & \textbf{Ratio} & \textbf{Mass range}
        & \textbf{Raw cos.} & \textbf{Zeroed cos.} \\
        \midrule
        Gemma-3-1B   & 1038 & $525\times$  & 0.2--\textbf{52.7}\%  & 0.540 & 0.642 \\
        \rowcolor{rowgray}
        Gemma-3-4B   & 443  & $2093\times$ & \textbf{85.5--92.6}\% & \textbf{0.960} & 0.643 \\
        Gemma-3-12B  & 2339 & $2198\times$ & \textbf{83.9--93.2}\% & \textbf{0.965} & 0.673 \\
        \rowcolor{rowgray}
        Qwen3-4B     & 0    & $54\times$   & 0.1--3.8\%   & 0.679 & 0.677 \\
        Qwen3.5-9B   & 3994 & $166\times$  & 0.0--1.3\%   & 0.667 & 0.668 \\
        \rowcolor{rowgray}
        Llama-3.1-8B & 4055 & $38\times$   & 0.1--1.9\%   & 0.658 & 0.657 \\
        \bottomrule
    \end{tabular}

    \caption{\textbf{Massive-activation dimensions and their effect on cross-lingual similarity.}
    \textit{Ratio} is the mean absolute magnitude of the highest-ranked dimension relative to the median dimension.
    \textit{Mass range} is the fraction of the direction's squared norm carried by that dimension across languages.
    \textit{Raw cos.} and \textit{Zeroed cos.} are the mean pairwise cross-lingual cosine similarities before and after zeroing the identified dimension.}
    \label{tab:massive_activation}
\end{table*}

\subsection{Massive-Activation Dimensions in Gemma Models}
\label{app:massive_activation}

During our analysis of the Gemma models, we encountered \textit{massive-activation dimensions} (MADs): individual residual-stream dimensions whose activation magnitudes are orders of magnitude larger than the others \citep{sun2024massive, timkey-van-schijndel-2021-bark}.
At the selected $(l,p)$ for each model, we find exactly one such dimension in each Gemma model, but no comparably dominant dimension in Qwen or Llama.

We find that these MADs dominate the norm of the extracted negation directions and substantially inflate the cross-lingual cosine similarity. 
However, they have no causal importance regarding negation polarity, as zeroing them out does not change the intervention results. 
We therefore remove their effects from similarity analyses, while leaving them unchanged during intervention experiments.
We leave a fuller investigation of the origin and broader effects of these massive-activation dimensions to future work.

\paragraph{Detection of massive-activation dimensions.}
At each model's selected $(l,p)$, we consider the four language-specific negation directions $\{{\mathbf d}^L\}$.
For each residual dimension $i$, we compute its mean absolute magnitude across languages,
\begin{equation}
    m_i = \frac{1}{|\mathcal{L}|}
    \sum_{L \in \mathcal{L}} |\mathbf {d}^L[i]|,
\end{equation}
and rank dimensions by $m_i$.
We inspect the highest-ranked dimension by comparing its magnitude to the median dimension and by computing the fraction of each direction's squared norm that it carries,
\begin{equation}
    \rho_L(i)
    =
    \frac{\mathbf {d}^L[i]^2}
         {\|\mathbf {d}^L\|_2^2}.
\end{equation}

Table~\ref{tab:massive_activation} shows that each Gemma model contains one extreme dimension.
For Gemma-3-4B and 12B, the identified dimension is over $2000\times$ larger than the median dimension and single-handedly accounts for $83.9$--$93.2\%$ of the squared norm across languages.
Gemma-3-1B also contains a large outlier, although its contribution is asymmetric across languages.
In contrast, the highest-ranked dimensions in Qwen and Llama account for at most $3.8\%$ of the direction norm.

\paragraph{Effects on cosine similarity.}
The MADs substantially inflate cross-language cosine similarity in Gemma-3-4B and 12B.
Before correction, their mean pairwise cross-lingual cosine similarities are $0.960$ and $0.965$, respectively.
After zeroing the single MAD, these decrease to $0.643$ and $0.673$.
Gemma-3-1B exhibits the opposite effect: its MAD contributes unevenly across languages, and removing it increases mean pairwise cosine from $0.540$ to $0.642$ (Table~\ref{tab:massive_activation}).
For Qwen and Llama, removing the highest-ranked dimension changes mean cosine by at most $0.002$.
Therefore, for all cosine similarity analyses involving Gemma models, we zero the identified MAD before computing cosine similarity.

\paragraph{Effects on causal interventions.}
We next test whether the same dimensions distort results from the intervention experiments.
For each Gemma model, we zero the identified MAD from the negation direction and rescale the direction to its original norm before performing within- and cross-language interventions (same setup as \S\ref{sec:crosslingual}).
We find that zeroing the MAD does not change intervention results: the mean absolute change in flip rate is $0.7$, $0.4$, and $0.5$ percentage points for Gemma-3-1B, 4B, and 12B, respectively; the largest changes are only $3$, $2$, and $4$ points (Table~\ref{tab:massive_activation_steering}).
Thus, we leave the MADs unchanged for intervention experiments throughout the paper.

\begin{table}[h]
    \centering
    \small
    \setlength{\tabcolsep}{5pt}
    \begin{tabular}{l cc}
        \toprule
        \rowcolor{headergray}
        \textbf{Model} & \textbf{Mean $|\Delta|$ flip} & \textbf{Max. $|\Delta|$ flip} \\
        \midrule
        Gemma-3-1B  & 0.7 pp & 3 pp \\
        \rowcolor{rowgray}
        Gemma-3-4B  & 0.4 pp & 2 pp \\
        Gemma-3-12B & 0.5 pp & 4 pp \\
        \bottomrule
    \end{tabular}

    \caption{\textbf{Effect of removing the massive-activation dimension on causal steering.}
    Changes are computed across the full cross-language transfer matrices for both addition and removal.}
    \label{tab:massive_activation_steering}
\end{table}

\subsection{Target Language Ablation}
\label{app:varying_target}

We test whether varying the target translation language would change the resulting extracted direction and consequently, the causal intervention results.

\paragraph{Setup.} For all six models, we keep the source language fixed as EN, while varying the target language to VI, FR and ZH.
For each translation pair, we extract the one negation direction at the same ($l,p$) as the EN$\rightarrow$VI direction in \S\ref{sec:linear}.
We first compare the cosine similarity of the directions, with Gemma family's directions having their MAD zeroed out. 
We then intervene with those directions on corresponding translation pair and measure the flip rates, using the same $\alpha$ as EN$\rightarrow$VI.

\paragraph{Results.}
Changing the target language has little effect on the extracted direction: across all six models, the pairwise cosine similarities range from $0.9896$ to $0.9986$ and remain near-identical across target languages (Table~\ref{tab:target_ablation}).
Steering results are similarly stable: relative to EN$\rightarrow$VI, changing the target produces a mean absolute difference of $5.1$ percentage points in flip rate across models, targets, and intervention modes. 
Thus, we used a single target language per source in the main experiments.

\begin{table}[h]
    \centering
    \small

     \begin{subtable}{\linewidth}
        \centering
        \setlength{\tabcolsep}{4.5pt}
        \begin{tabular}{l ccc}
            \toprule
            \rowcolor{headergray}
            \textbf{Model} & \textbf{VI--FR} & \textbf{VI--ZH} & \textbf{FR--ZH} \\
            \midrule
            Gemma-3-1B   & 0.9973 & 0.9980 & 0.9942 \\
            \rowcolor{rowgray}
            Gemma-3-4B   & 0.9953 & 0.9957 & 0.9912 \\
            Gemma-3-12B  & 0.9930 & 0.9945 & 0.9896 \\
            \rowcolor{rowgray}
            Qwen3-4B     & 0.9961 & 0.9975 & 0.9949 \\
            Qwen3.5-9B   & 0.9969 & 0.9986 & 0.9961 \\
            \rowcolor{rowgray}
            Llama-3.1-8B & 0.9948 & 0.9956 & 0.9901 \\
            \bottomrule
        \end{tabular}
        \caption{\textbf{Pairwise cosine similarity.}
        For Gemma models, cosine similarity is computed after zeroing the identified massive-activation dimension.
        }
    \end{subtable}
    
    \vspace{0.7em}

\begin{subtable}{\linewidth}
    \centering
    \setlength{\tabcolsep}{4.2pt}
    \begin{tabular}{l ccc}
        \toprule
        \rowcolor{headergray}
        \textbf{Model} & \textbf{EN$\rightarrow$VI} & \textbf{EN$\rightarrow$FR} & \textbf{EN$\rightarrow$ZH} \\
        \midrule
        Gemma-3-1B   & 86.9 / 77.0 & 96.0 / 87.9 & 79.0 / 88.4 \\
        \rowcolor{rowgray}
        Gemma-3-4B   & 98.0 / 88.0 & 99.0 / 91.9 & 93.9 / 85.9 \\
        Gemma-3-12B  & 99.0 / 95.0 & 100.0 / 97.0 & 89.9 / 91.6 \\
        \rowcolor{rowgray}
        Qwen3-4B     & 90.9 / 92.7 & 99.0 / 97.0 & 92.9 / 91.9 \\
        Qwen3.5-9B   & 98.0 / 93.0 & 100.0 / 99.0 & 90.9 / 87.8 \\
        \rowcolor{rowgray}
        Llama-3.1-8B & 94.9 / 91.0 & 97.0 / 98.0 & 87.9 / 85.9 \\
        \bottomrule
    \end{tabular}
    \caption{\textbf{Addition / removal rates (\%).}
    }
\end{subtable}

    \caption{\textbf{Target-language ablation with EN as the fixed source language.}
    Top: pairwise cosine similarity between negation directions extracted with VI, FR, and ZH as translation targets.
    Bottom: addition and removal rates using the corresponding target-specific direction.}
    \label{tab:target_ablation}
\end{table}


\subsection{Random-Direction Control}
\label{app:random_control}

To test whether any perturbation of the same magnitude as $\mathbf{d}_\text{CAA}$ can flip negation with the same success, we use random directions for intervention.

\paragraph{Setup.} We sample $\mathbf{d}_\text{random}\sim\mathcal{N}(\mathbf{0},\mathbf{I}_{d_\text{model}})$ (using $\texttt{random\_seed}\in[0,4]$) and rescale it to $\lVert\mathbf{d}^{L}_{\text{CAA}}\rVert$. We intervene with this vector at each model's selected $(l,p)$ with the same $\alpha$ as \S\ref{sec:linear}. 
The random directions are nearly orthogonal to the negation directions ($|\cos(\mathbf{d}_\text{random}, \mathbf{d}_\text{CAA})| \leq$ 0.06). 
Table~\ref{tab:random_control} reports the median flip rate across seeds.

\begin{table}[h]
    \centering
    \small
    \setlength{\tabcolsep}{3.3pt}
    \begin{tabular}{l cccc}
        \toprule
        \rowcolor{headergray}
        \textbf{Model} & \textbf{EN} & \textbf{VI} & \textbf{FR} & \textbf{ZH} \\
        \midrule
        Gemma-3-1B   & 0 / 0 & 0 / 2 & 0 / 1 & 0 / 8 \\
        \rowcolor{rowgray}
        Gemma-3-4B   & 0 / 1 & 0 / 6 & 0 / 0 & 0 / 6 \\
        Gemma-3-12B  & 0 / 4 & 0 / 2 & 0 / 5 & 0 / 5 \\
        \rowcolor{rowgray}
        Qwen3-4B     & 0 / 0 & 0 / 1 & 0 / 0 & 0 / 1 \\
        Qwen3.5-9B   & 0 / 0 & 0 / 0 & 0 / 0 & 0 / 1 \\
        \rowcolor{rowgray}
        Llama-3.1-8B & 0 / 0 & 0 / 0 & 0 / 0 & 0 / 1 \\
        \bottomrule
    \end{tabular}

    \caption{\textbf{Random-direction control: addition / removal rates (\%).}
    Median over 5 seeds of norm-matched isotropic random directions, applied
    with the same $(l,p)$ and $\alpha$ as \S\ref{sec:linear}.}
    \label{tab:random_control}
\end{table}

\paragraph{Results.} Random directions almost never induce polarity flips. Addition rates are 0\% for every model and language, while removal rates remain low, reaching at most 8\% (Gemma-3-1B, ZH). 
In contrast, CAA steering yields substantially higher flip rates across all models, languages, and directions (Table~\ref{tab:direction_peaks}). 
The asymmetry suggests that random perturbation to the source can cause models to drop the negation markers, but not insert them.

\section{Reproducibility Details}
\label{app:reproducibility}

\subsection{Implementation Details}
\paragraph{Inference settings.} All generations use greedy decoding with \texttt {max\_new\_tokens = 64 \text{and} 128} for Mini-minus-4 and FLORES-minus-4 respectively, and \texttt{do\_sample=False}. For Qwen models, we disable thinking mode with \texttt{enable\_thinking=False}.

\paragraph{Sparse autoencoders.}
\label{app:sae-specs}

We use publicly released sparse autoencoders (SAEs) for all three model families.
Table~\ref{tab:sae-specs} summarises the key properties of each release.

\begin{table*}[h]
\centering
\small
\begin{threeparttable}
\begin{tabular}{llllll}
\toprule
\textbf{Training Activations} & \textbf{SAE Release} & \textbf{Activation} & \textbf{Width} & \textbf{Site} & \textbf{L0} \\
\midrule
Gemma-3-4B-IT & Gemma Scope 2\footnotemark[1] & JumpReLU & 16k & Residual stream & ``big" (90-150) \\
Qwen3.5-9B-Base & Qwen SAE-Res\footnotemark[2] & TopK & 64k & Residual stream & 100 \\
Llama-3.1-8B-Base & Llama Scope\footnotemark[3] & JumpReLU & 32k (8$\times$) & Residual stream & 50 \\
\bottomrule
\end{tabular}
\begin{tablenotes}
\item[1] \url{https://huggingface.co/google/gemma-scope-2-4b-it/tree/main/resid_post_all}
\item[2] \url{https://huggingface.co/Qwen/SAE-Res-Qwen3.5-9B-Base-W64K-L0_100}
\item[3] \url{https://huggingface.co/OpenMOSS-Team/Llama3_1-8B-Base-LXR-8x}
\end{tablenotes}
\caption{Sparse autoencoder releases used in this work. Width = dictionary size; L0 = mean number of active features per token at convergence.}
\label{tab:sae-specs}
\end{threeparttable}
\end{table*}

\paragraph{Base vs.\ instruct SAEs.}
Gemma Scope 2 provides SAEs trained directly on the instruct-tuned model, matching our experimental target.
For Llama Scope and Qwen, only base-model SAEs are available; we apply these to the instruct-tuned variants.
Both releases report strong reconstruction transfer from base to instruct checkpoints, and our causal results (feature-level steering flip rates) empirically confirm that the dictionaries remain effective on the instruct models.

\paragraph{Compute.} All experiments are conducted on 1$\times$ A100 40GB.

\subsection{Evaluating Flip} 
\label{app:steering-eval}
\subsubsection{Using Pre-determined Rules}
We provide the thresholds for a successful intervention on Mini-minus-4 dataset below. 
\begin{itemize}
    \item \textbf{Language identification}: fastText's (\texttt{lid.176.bin}\footnote{https://fasttext.cc/docs/en/language-identification}) top-1 prediction is the intended language and confidence is $\ge$0.5.
    \item \textbf{Non-degeneracy}: The output sentence contains 2--200 words and $\ge$30\% of tokens are unique.
    \item \textbf{Content preservation}: For Mini-minus-4, the output scores $\ge$30 chrF against the model's un-intervened translation of the opposite polarity. 
\end{itemize}

\subsubsection{Using LLM-as-Judge}

\begin{figure*}[t]
\centering
\begin{tcolorbox}[colback=gray!5!white, colframe=gray!60!black, boxrule=0.5mm, arc=2mm]
\small

\noindent\textbf{System message:}
\begin{quote}
\small
You are a strict linguistic annotator comparing two translations of the same source.
Answer only in JSON.
Judge meaning, not surface form.
You do NOT know which translation received any intervention; report differences neutrally.
IMPORTANT: a polarity flip alone (B says the opposite of A) is NOT a meaning-preservation failure --- polarity is reported separately in questions 1 and 2.
\\
\end{quote}

\noindent\textbf{User message:}
\begin{quote}
\small
You compare two translations of the same sentence and report polarity + meaning.

\medskip
\noindent
Source (\{source\_lang\}): \{source\} \\
Main clause of the source: \{main\_clause\} \\
Translation A (\{target\_lang\}): \{baseline\} \\
Translation B (\{target\_lang\}): \{intervened\}

\medskip
\noindent
Answer three questions about the two translations.
\end{quote}

\begin{enumerate}[nosep,leftmargin=2.4em]
    \item \texttt{main\_clause\_polarity\_A}: polarity of the main clause in A. \\
          \texttt{"negated" | "affirmative" | "missing" | "unclear"}

    \item \texttt{main\_clause\_polarity\_B}: polarity of the main clause in B. \\
          \texttt{"negated" | "affirmative" | "missing" | "unclear"}

    \item \texttt{meaning\_preserved}: if you forced A and B to have the
          \emph{same} polarity, would they describe the same event/situation
          with the same participants, actions, objects, and modifiers?
          This is about topic and structure, \emph{not} truth value.
          A negated version of A that keeps the same participants/action is
          \texttt{"yes"} here; do not lower this answer just because polarity changed. \\
          \texttt{"yes"} --- same event, same participants/actions/objects
          (polarity may differ) \\
          \texttt{"no"} --- B is off-topic, gibberish, a refusal, wrong language,
          or describes a different event entirely
\end{enumerate}

\begin{quote}
\small
\noindent Rules:
\begin{itemize}[nosep,leftmargin=1.2em]
    \item Judge meaning, not surface words. Any grammatical negation form counts
          (không, ne\ldots pas, 不/没, prefix \emph{un-}, etc.).
    \item Do not reward or penalise polarity itself; just report it.
    \item Questions 1, 2 and 3 are independent. A ``flipped'' polarity must not
          lower \texttt{meaning\_preserved} by itself.
\end{itemize}

\medskip
\noindent
Answer in JSON only:
\begin{verbatim}
{
  "main_clause_polarity_A": "...",
  "main_clause_polarity_B": "...",
  "meaning_preserved": "...",
  "evidence": "<<=15 words>"
}
\end{verbatim}
\end{quote}

\end{tcolorbox}

\caption{\textbf{LLM-as-judge prompt for evaluating steering interventions on FLORES-minus-4}.
}
\label{fig:flip_judge_prompt}
\end{figure*}

On FLORES-minus-4, we replace the heuristic polarity rules used for Mini-minus-4 with an LLM-as-judge protocol (GPT-5.6-Terra) using the prompt in Figure~\ref{fig:flip_judge_prompt}.

\paragraph{Baseline eligibility.}
For each item, we first determine whether the un-intervened baseline translation preserves the source polarity. Given the source sentence, the annotated main clause, and the baseline output, the judge classifies the polarity of the translated main clause as \textit{negated}, \textit{affirmative}, \textit{missing}, or \textit{unclear}. A row is \textbf{eligible} for scoring only if the baseline polarity matches the source polarity; otherwise, the model did not preserve the source polarity before intervention, so the effect cannot be cleanly attributed to the intervention.

\paragraph{Intervention evaluation.}
For eligible items, the judge evaluates the baseline and intervened translations as a blind pair, labeled A and B, without being told which output received the intervention. 
Given the source, annotated main clause, and both translations, the judge reports three fields: (1)~the main-clause polarity of A, (2)~the main-clause polarity of B, and (3)~whether the two translations preserve the same underlying event or situation when polarity differences are set aside.

\paragraph{Flip rate.}
The \textbf{flip rate} is the proportion of eligible items for which the intervention successfully changes the main-clause polarity to the target polarity while retaining meaning.

\subsection{Human Validation}
\label{app:human_validation_judge}

\paragraph{Protocol.}
We validate the reliability of the automated Flip Rate against human evaluation. 
We employ three human annotators to independently evaluate  Gemma-3-4B's candidate flips on both Mini-minus-4 (100 samples $\times$ 2 flip directions for each language) and FLORES-minus-4 (216 $\times$ 2 for each language), covering all four languages and both intervention modes. One annotator evaluates both EN$\xrightarrow{}$VI and VI$\xrightarrow{}$EN.

\begin{figure*}
    \centering
    \includegraphics[width=1\linewidth]{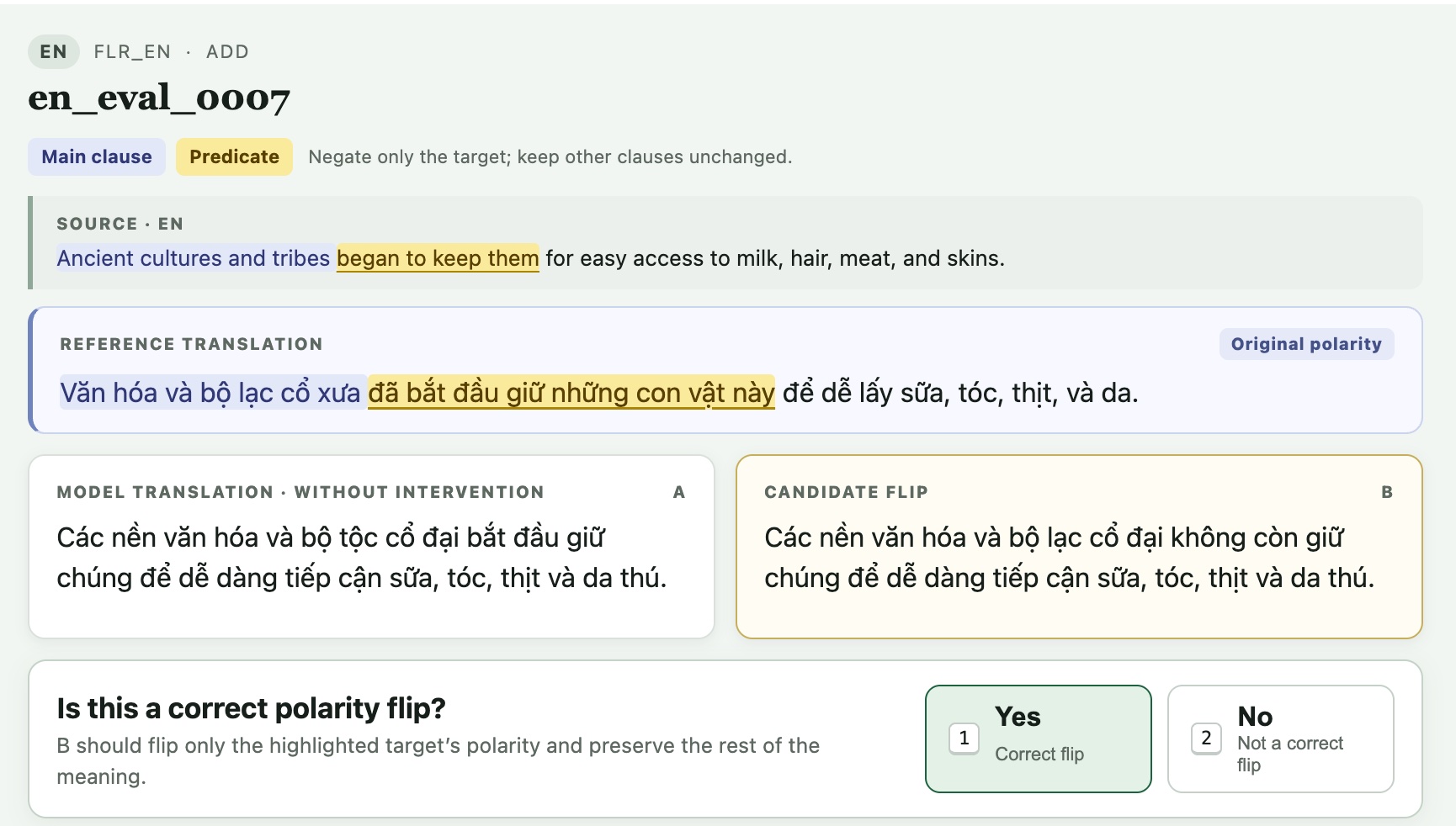}
    \caption{\textbf{Example of the annotation interface.} Human evaluators are shown one example at a time, including the source sentence, the reference translation and the un-intervened baseline, and the intervened output. The evaluator is asked to judge if the intervened output is a correct polarity flip, with meaning preserved otherwise.}
    \label{fig:annotation_example}
\end{figure*}

We build an annotation interface where the human evaluators are shown one example at a time: the source sentence, the reference translation and the (un-intervened) baseline, and the intervened output. Example of the annotation screen is shown in Figure~\ref{fig:annotation_example}.
The evaluators are asked to judge if the intervened version is a polarity flip of the baseline with no content drift. They do not have access to the automated evaluation.

\paragraph{Results.}
We measure agreement between human judgments and the rule-based evaluator (Mini-minus-4) or LLM-as-judge (FLORES-minus-4). 
We report agreement percentage, Cohen's $\kappa$ \cite{cohen_kappa}, prevalence-adjusted bias-adjusted kappa (PABAK) \cite{byrt1993bias}, and precision and recall for identifying successful interventions (Table~\ref{tab:human_validation}). 
Because the evaluations are heavily skewed toward success, which can decrease Cohen's $\kappa$ despite high observed agreement, we additionally report PABAK to account for the imbalanced data.

On Mini-minus-4, the rule-based evaluator achieves 96.5\% agreement ($\kappa=0.732$, PABAK $=0.930$). Only one false positive and 27 false negatives are observed across 800 examples. 
On FLORES-minus-4, the LLM judge achieves 93.7\% agreement ($\kappa=0.612$, PABAK $=0.874$) in 1,728 examples. 
Overall, the high agreement rates support the reliability of our automatic evaluation methods, although FLORES-minus-4 flip rates may be subject to upward bias.

\begin{table}[t]
\centering
\footnotesize
\setlength{\tabcolsep}{2pt}
\renewcommand{\arraystretch}{1.22}

\begin{subtable}{\linewidth}
\centering
\begin{tabular}{@{}l r r r r r r r r@{}}
\toprule
\rowcolor{headergray}
\textbf{Lang.} & $\boldsymbol{n}$ &
\textbf{Agree.} & \textbf{PABAK} &
$\boldsymbol{\kappa}$ &
\textbf{FP} & \textbf{FN} &
\textbf{Prec.} & \textbf{Rec.} \\
\midrule

EN & 200 & 98.5 & 0.970 & 0.901 & 0 & 3 & 100.0 & 98.4 \\

\rowcolor{rowgray}
VI & 200 & 92.5 & 0.850 & 0.643 & 0 & 15 & 100.0 & 91.8 \\

FR & 200 & 97.0 & 0.940 & 0.735 & 1 & 5 & 99.5 & 97.4 \\

\rowcolor{rowgray}
ZH & 200 & 98.0 & 0.960 & 0.492 & 0 & 4 & 100.0 & 98.0 \\

\midrule
\textbf{All} & \textbf{800} &
\textbf{96.5} & \textbf{0.930} &
\textbf{0.732} & \textbf{1} & \textbf{27} &
\textbf{99.9} & \textbf{96.4} \\

\bottomrule
\end{tabular}
\caption{\textbf{Mini-minus-4.}}
\label{tab:human_validation_mini}
\end{subtable}

\vspace{4pt}

\begin{subtable}{\linewidth}
\centering
\begin{tabular}{@{}l r r r r r r r r@{}}
\toprule
\rowcolor{headergray}
\textbf{Lang.} & $\boldsymbol{n}$ &
\textbf{Agree.} & \textbf{PABAK} &
$\boldsymbol{\kappa}$ &
\textbf{FP} & \textbf{FN} &
\textbf{Prec.} & \textbf{Rec.} \\
\midrule

EN & 432 & 92.1 & 0.843 & 0.549 & 33 & 1 & 91.9 & 99.7 \\

\rowcolor{rowgray}
VI & 432 & 94.4 & 0.889 & 0.664 & 22 & 2 & 94.5 & 99.5 \\

FR & 432 & 94.4 & 0.889 & 0.588 & 24 & 0 & 94.2 & 100.0 \\

\rowcolor{rowgray}
ZH & 432 & 93.8 & 0.875 & 0.644 & 27 & 0 & 93.3 & 100.0 \\

\midrule
\textbf{All} & \textbf{1728} &
\textbf{93.7} & \textbf{0.874} &
\textbf{0.612} & \textbf{106} & \textbf{3} &
\textbf{93.5} & \textbf{99.8} \\

\bottomrule
\end{tabular}
\caption{\textbf{FLORES-minus-4.}}
\label{tab:human_validation_flores}
\end{subtable}

\caption{\textbf{Human validation of automatic intervention evaluation.}
Agreement (Agree.), precision (Prec.), and recall (Rec.) are reported in percentages. FP (False Positives) and FN (False Negatives) denote automatic evaluation errors relative to human judgments.}
\label{tab:human_validation}
\end{table}

\section{Details of Dataset Construction}
\label{app:data-construction}

\begin{table*}[t]
\centering
\small
\renewcommand{\arraystretch}{1.12}
\setlength{\tabcolsep}{5pt}

\begin{tabularx}{\textwidth}{
    @{} l
    >{\raggedright\arraybackslash}X
    >{\raggedright\arraybackslash}X @{}
}
\toprule
\multicolumn{3}{c}{\textbf{Mini-minus-4}} \\
\midrule
\textbf{Lang.} & \textbf{Affirmative} & \textbf{Negative} \\
\midrule
EN &
She \textcolor{affcolor}{\textbf{is}} at the store. &
She \textcolor{negcolor}{\textbf{is not}} at the store. \\

VI &
Tôi \textcolor{affcolor}{\textbf{thích}} phở. &
Tôi \textcolor{negcolor}{\textbf{không thích}} phở. \\

FR &
Nous \textcolor{affcolor}{\textbf{aimons}} cette chanson. &
Nous \textcolor{negcolor}{\textbf{n'aimons pas}} cette chanson. \\

ZH &
他\textcolor{affcolor}{\textbf{喜欢}}跑步。 &
他\textcolor{negcolor}{\textbf{不喜欢}}跑步。 \\

\midrule
\multicolumn{3}{c}{\textbf{FLORES-minus-4}} \\
\midrule
EN &
Schneider \textcolor{affcolor}{\textbf{testified}} via videolink from a USAF base in his homeland. &
Schneider \textcolor{negcolor}{\textbf{did not testify}} via videolink from a USAF base in his homeland. \\

VI &
Schneider \textcolor{affcolor}{\textbf{làm chứng}} qua kết nối video từ căn cứ USAF ở quê nhà. &
Schneider \textcolor{negcolor}{\textbf{không làm chứng}} qua kết nối video từ căn cứ USAF ở quê nhà. \\

FR &
Schneider \textcolor{affcolor}{\textbf{a témoigné}} par liaison vidéo depuis une base de l'USAF dans son pays. &
Schneider \textcolor{negcolor}{\textbf{n'a pas témoigné}} par liaison vidéo depuis une base de l'USAF dans son pays. \\

ZH &
施耐德 (Schneider) 在其祖国的美国空军 (USAF) 基地通过视频连线
\textcolor{affcolor}{\textbf{进行作证}}。 &
施耐德 (Schneider) 在其祖国的美国空军 (USAF) 基地
\textcolor{negcolor}{\textbf{没有}}通过视频连线进行作证。 \\

\bottomrule
\end{tabularx}

\caption{\textbf{Examples from Mini-minus-4 and FLORES-minus-4.}
Each row shows an affirmative/negative contrast in one of the four languages.
Mini-minus-4 consists of controlled minimal pairs, whereas FLORES-minus-4 contains polarity contrasts constructed from parallel FLORES-200 sentences.}
\label{tab:dataset_examples}
\end{table*}

\subsection{Mini-minus-4}
\label{app:mini-minus-4}

Mini-minus-4\footnote{``Mini" indicates this is a dataset of minimally contrastive sentence pairs and the sentences themselves are short and minimalistic.} is a synthetic minimal-pair dataset covering English (EN), Vietnamese (VI), French (FR), and Chinese (ZH). 
For each language, we generate single-clause sentences paired with their polarity-flipped counterparts, differing only by the presence of a negation marker. Each language is pinned to a single negation regime to avoid confounds:

\begin{itemize}
    \item \textbf{EN}: copula or modal frame (\textit{be} + adjective, or \textit{can/will/should} + verb), negated with \textit{not}. This avoids do-support, ensuring the only difference between pairs is the negator.
    \item \textbf{VI}: preverbal \textit{không} only, excluding aspect- or register-carrying alternatives (\textit{chưa}, \textit{chẳng}, \textit{chả}, \textit{đâu}, \textit{đừng}).
    \item \textbf{FR}: standard written \textit{ne\ldots pas} with elision, formal register. Excludes \textit{jamais}, \textit{plus}, \textit{rien}, \textit{personne}, \textit{aucun}, and expletive \textit{ne}.
    \item \textbf{ZH}: 不 with stative, adjectival, habitual, or volitional predicates. Excludes 没/没有/别/未, which carry aspect or mood distinctions.
\end{itemize}

\paragraph{Generation.} Pairs are generated natively in each language using GPT-5.4-mini (temperature 0.9), with prompts rotated across 20 everyday domains (food, travel, health, sports, etc.) for lexical diversity. Each prompt requires that the two sentences differ only by the negation marker, with identical subject, predicate, and tense. We overgenerate by a factor of 1.3$\times$ to allow for filtering losses.

\paragraph{Filtering.} Generated pairs pass through three filters: (1)~per-language marker validation, which verifies that the negative sentence contains exactly the expected marker in the correct position and that no other negation markers are present; (2)~deduplication by sentence hash and predicate lemma; and (3)~a predicate-disjoint split (seed 42), which partitions pairs into 150 for extraction, 100 for evaluation and 100 for test, guaranteeing that no predicate appears in more than one split.

\paragraph{Final yield.} 350 pairs per main language (150 extraction + 100 evaluation + 100 test).

\paragraph{Additional-language test set.}
We additionally generate 100 held-out test pairs each in Russian (RU), Hindi (HI), Arabic (AR), and Indonesian (ID), using the same minimal-pair construction. 
Each language is restricted to a single verbal negation regime:
preverbal {\russianfont не} in RU,
preverbal {\hindifont नहीं} in HI,
preverbal {\arabicfont لا} with imperfect verbs in AR,
and preverbal \textit{tidak} in ID.

\subsection{FLORES-minus-4}
\label{app:flores-minus-4}

We derive FLORES-minus-4\footnote{The ``minus'' indicates that the dataset is a negated version (since the original dataset is called FLORES+), while 4 indicates the number of languages.} from the FLORES-200 devtest set~\citep{nllb2024scaling} through a three-stage pipeline: candidate collection, LLM-assisted flip generation, and LLM verification. 

\paragraph{Stage 1: Candidate collection.}
We begin with the 1{,}012-sentence devtest split, restricted to the parallel intersection of our four languages (EN, VI, FR, ZH). Sentences are shuffled with a fixed seed to avoid the domain clustering present in the original row order. We retain sentences with 6--30 English words and partition them by polarity. The \textbf{positive pool} excludes sentences containing any English negation marker, including morphological forms (\textit{un-}, \textit{non-}, \textit{-less}). The \textbf{negative pool} retains only sentences with verbal analytic negation (\textit{not}, \textit{n't}), excluding lexical negators (\textit{never}, \textit{nothing}) and morphological negation. 

Both pools reject sentences with neg-raising verbs (\textit{think}, \textit{believe}, \textit{seem}) and sentences whose parallel Vietnamese or Chinese translations contain non-standard negators (e.g., VI \textit{chưa}/\textit{chẳng}, ZH 未能/从未). This yields \textbf{795 candidate} parallel 4-tuples.

GPT-5.6-Luna also annotates the main clause where polarity differs for each language, returning verbatim substrings used to scope the polarity judge and define intervention token spans. 

\paragraph{Stage 2: Flip generation and annotation.}
For each candidate, GPT-5.6-Luna generates a polarity-flipped counterpart in all four languages. 
The prompt enforces the following constraints: use only standard verbal negation (EN \textit{did not}/\textit{does not}; VI \textit{không}; FR \textit{ne\ldots pas}; ZH 不/没有 selected by aspect); preserve all content words; apply only grammar-forced co-changes (e.g., EN \textit{some}$\leftrightarrow$\textit{any}, FR partitive \textit{du}$\rightarrow$\textit{de}); and reject rather than produce a strained flip. 
The model outputs structured JSON that records the target predicate, aspect, and negator choice before the flipped sentence, preventing post-hoc rationalization. This produces \textbf{439 items} with at least one successful flip.

\paragraph{Stage 3: Verification.}
A verifier (GPT-5.6-Terra) judges each item. Per-language checks verify that: (1)~the flip reverses propositional polarity, (2)~negation targets the main predicate rather than a subordinate clause, (3)~modal scope is correctly realized, (4)~no content beyond grammar-forced changes was altered, (5)~the negator suits the aspect and predicate type, and (6)~the flip is natural. Item-level checks verify that (7)~all four languages negate the same proposition and (8)~the negative sentence does not introduce a scope ambiguity that would translate differently across languages. Items where the verifier is uncertain are flagged for human review rather than silently dropped.

\paragraph{Final dataset.}
The final FLORES-minus-4 set consists of items that pass all verification checks and have valid main-clause annotations in all four languages, yielding \textbf{216 parallel sentences}.

\section{Extensions of Main Results}

\subsection{Negation Translation Evaluation on FLORES-minus-4}
\label{app:preservation}

We evaluate baseline (un-intervened) translations on FLORES-minus-4 using four complementary metrics.
Neural and n-gram metrics are widely reported to misjudge whether the output translations maintain correct source polarity \cite{tang-etal-2021-revisiting}, therefore we rely on them solely to evaluate the overall translation quality. We use LLM-as-judge as the main evaluator of whether negation is accurately preserved.
 
\paragraph{Polarity Preservation (\%) --- LLM-as-Judge.}
For each (source, hypothesis) pair, an LLM judge (GPT-5.6-Terra) labels the main-clause polarity of the hypothesis as \textit{affirmative} or \textit{negated}, using the prompt in Figure~\ref{fig:polarity_annotation_prompt}.
A row counts as \emph{preserved} when the judge's label matches the source's ground-truth polarity.
Rows labeled \textit{missing} (main clause not translated, refusal, commentary, or wrong language) or \textit{unclear} (ambiguous negation scope) count as non-preservation.

\begin{figure*}[t]
\centering
\begin{tcolorbox}[colback=gray!5!white, colframe=gray!60!black, boxrule=0.5mm, arc=2mm]
\small
\noindent\textbf{System message:}
\begin{quote}
\small
You are a strict linguistic annotator. Answer only in JSON.
\end{quote}

\noindent\textbf{User message:}
\begin{quote}
\small
You judge whether a translation keeps the polarity of one specific clause.
\medskip
\noindent
Source (\{source\_lang\}): \{source\} \\
Main clause of the source: \{main\_clause\} \\
Main clause polarity in source: \{source\_polarity\} \hfill \texttt{\# "negated" or "affirmative"} \\
Translation (\{target\_lang\}): \{hypothesis\}

\medskip
\noindent
Question: In the translation, what is the polarity of the part that translates the main clause?
\end{quote}

\noindent Rules:
\begin{itemize}[nosep,leftmargin=2.4em]
    \item ``negated'': the main clause's meaning is negated (any grammatical form counts: không, ne\ldots pas, 不/没, etc.).
    \item ``affirmative'': the main clause's meaning is stated without negation.
    \item ``missing'': the main clause is not translated at all, or the output is not a translation (refusal, commentary, wrong language).
    \item ``unclear'': you cannot tell (e.g.\ the negation is present but its scope is ambiguous).
\end{itemize}

\begin{quote}
\small
Ignore all other clauses. \\
Ignore translation quality unless it makes the clause meaning unrecoverable.

\medskip
\noindent
Answer in JSON only: \\
\texttt{\{"main\_clause\_polarity": "negated|affirmative|missing|unclear",} \\
\texttt{~"evidence": "<the words in the translation that show it, <=10 words>"\}}
\end{quote}

\vspace{0.5em}
\end{tcolorbox}
\caption{\textbf{LLM-as-judge prompt for evaluating whether translation preserves the polarity of the source main clause.}}
\label{fig:polarity_annotation_prompt}
\end{figure*}

 
\paragraph{COMET-Kiwi (Reference-free QE) and COMET (Reference-based).}
We use \texttt{Unbabel/wmt22-cometkiwi-da} and \texttt{Unbabel/wmt22-comet-da} \citep{rei-etal-2022-cometkiwi, rei-etal-2022-comet}. COMET-Kiwi is a reference-free quality estimation model that takes only (source, hypothesis), while COMET is reference-based and takes (source, reference, hypothesis). Both models output a scalar in $[0,1]$.
 
\paragraph{chrF.}
We compute chrF via SacreBLEU \citep{popovic-2015-chrf}: a reference-based character $n$-gram F-score. Scores range from 0 to 100.

\begin{figure*}
    \centering
    \includegraphics[width=1\linewidth]{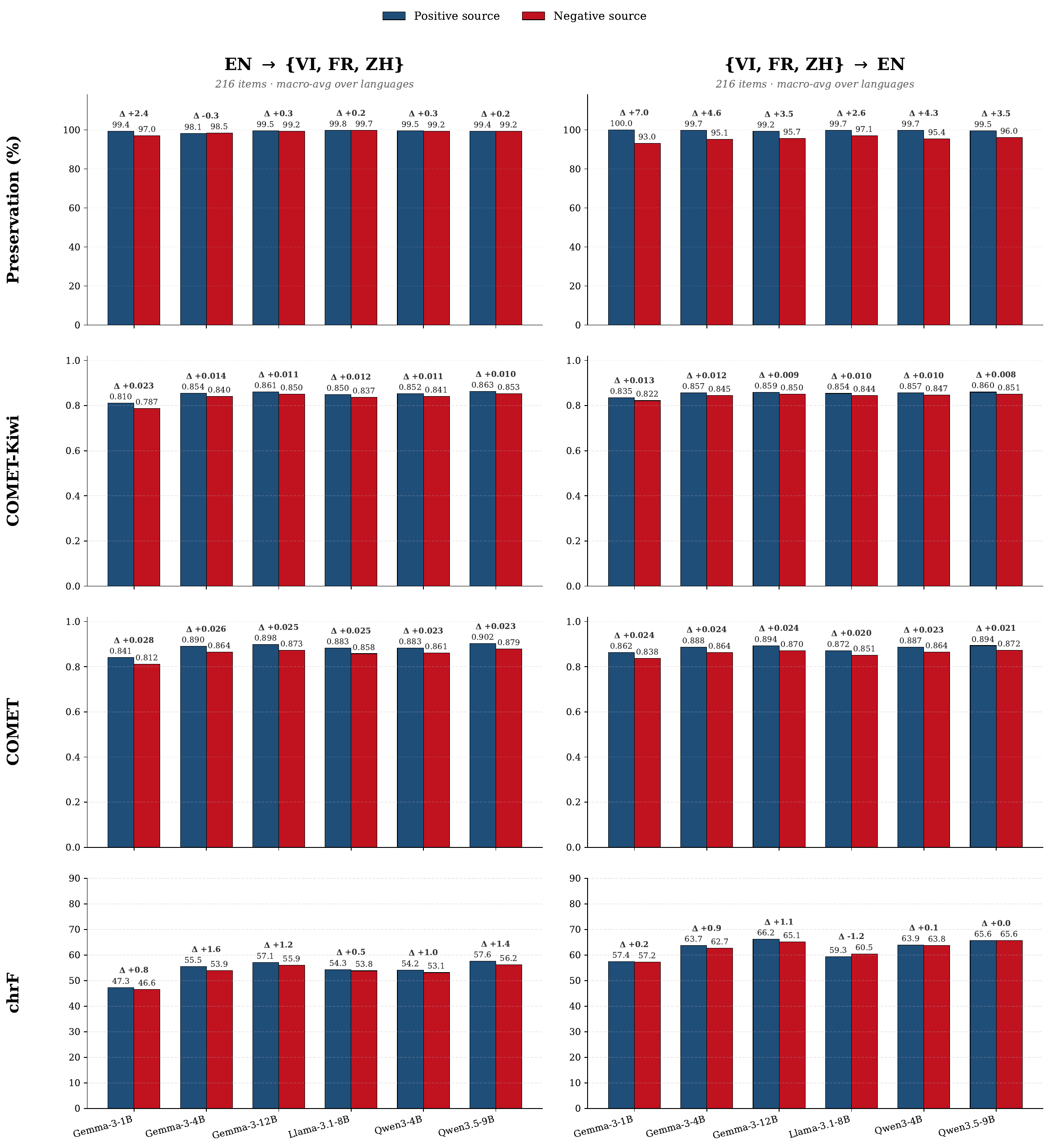}
    \caption{\textbf{Baseline translation evaluation on FLORES-minus-4.}
\textbf{Top}: Negation preservation (\%) via LLM-as-judge.
\textbf{Rows 2--4}: Translation quality measured by COMET-Kiwi (reference-free), COMET (reference-based), and chrF.
All six models preserve negation at ${\geq}97\%$ from English and ${\geq}93\%$ into English, with no substantial quality gap between polarities.}
    \label{fig:mt_baselines_cameraready}
\end{figure*}

\paragraph{Results.}
For each (model, direction) row, we average over all 216 items $\times$ 3 target languages, and report the results in Figure~\ref{fig:mt_baselines_cameraready}. Overall, we find that models preserve negation accurately, even though preservation for negated sentences is slightly lower than for positive sentences. The fluency of translations is also largely unaffected by the presence of negation in the source.

\subsection{Negation Preservation and Eligible Count on Mini-minus-4}
\label{app:eligible_count}

For our flip rate calculation, we compute the number of successful flips, divided by the number of eligible baselines. A baseline translation is considered eligible if its negation polarity is preserved from the source sentence (i.e., a negative sentence is correctly translated as negative). 
This is to avoid crediting the intervention with addition/removal of negation due to the model's translation failure. 


\textbf{Results.} On Mini-minus-4, we use deterministic filters for negation markers to determine if a baseline's negation polarity is negative or positive.
We report the number of baselines  which correctly preserved negation in Table~\ref{tab:eligible_count}. 
Overall, all models preserve negation at near perfect accuracy for most languages, with ZH negative examples the only exception.

\begin{table}[h]
\centering
\small

\begin{subtable}{\linewidth}
\centering
\setlength{\tabcolsep}{3.3pt}
\begin{tabular}{l cccc}
\toprule
\textbf{Model} & \textbf{EN} & \textbf{VI} & \textbf{FR} & \textbf{ZH} \\
\midrule
Gemma-3-1B   & 97 / 100 & 100 / 99  & 100 / 99  & 100 / 89 \\
Gemma-3-4B   & 98 / 100 & 100 / 98  & 100 / 100 & 100 / 94 \\
Gemma-3-12B  & 98 / 100 & 100 / 100 & 100 / 100 & 100 / 95 \\
Qwen3-4B     & 97 / 97  & 100 / 93  & 100 / 100 & 100 / 88 \\
Qwen3.5-9B   & 98 / 100 & 100 / 100 & 100 / 100 & 100 / 89 \\
Llama-3.1-8B & 98 / 100 & 100 / 97  & 100 / 100 & 100 / 96 \\
\bottomrule
\end{tabular}
\caption{Eval set. Counts are positive / negative samples.}
\label{tab:eligible_count_eval}
\end{subtable}

\vspace{4pt}

\begin{subtable}{\linewidth}
\centering
\setlength{\tabcolsep}{3.3pt}
\begin{tabular}{l cccc}
\toprule
\textbf{Model} & \textbf{EN} & \textbf{VI} & \textbf{FR} & \textbf{ZH} \\
\midrule
Gemma-3-1B   & 99 / 100 & 100 / 99  & 100 / 99  & 100 / 89 \\
Gemma-3-4B   & 99 / 100 & 100 / 97  & 100 / 100 & 100 / 97 \\
Gemma-3-12B  & 99 / 100 & 100 / 99  & 100 / 100 & 100 / 97 \\
Qwen3-4B     & 99 / 96  & 100 / 98  & 100 / 100 & 100 / 94 \\
Qwen3.5-9B   & 99 / 100 & 100 / 99  & 100 / 100 & 100 / 91 \\
Llama-3.1-8B & 99 / 100 & 100 / 98  & 100 / 100 & 100 / 96 \\
\bottomrule
\end{tabular}
\caption{Test set. Counts are positive / negative samples.}
\label{tab:eligible_count_test}
\end{subtable}

\caption{\textbf{Number of baseline translations with correctly preserved polarity.}}
\label{tab:eligible_count}
\end{table}

\begin{table}[t]
\centering
\renewcommand{\arraystretch}{1.1}
\begin{tabularx}{\linewidth}{@{} >{\raggedright\arraybackslash}X @{}}
\toprule
\textbf{Negation Removal} (ZH$\to$EN) \\
\midrule
\tagbox{SRC} 这台投影仪\textcolor{negcolor}{\textbf{不清楚}}。 \gloss{(unclear)} \\
\tagbox{BASE} This projector is \textcolor{negcolor}{\textbf{unclear}}. \\
\intbox\ This projector is \textcolor{affcolor}{\textbf{clear}}. \\
\midrule
\tagbox{SRC} 我们\textcolor{negcolor}{\textbf{不愿意}}参加讨论。 \gloss{(unwilling)} \\
\tagbox{BASE} We are \textcolor{negcolor}{\textbf{unwilling}} to participate in the discussion. \\
\intbox\ We \textcolor{affcolor}{\textbf{want}} to participate in the discussion. \\
\midrule
\tagbox{SRC} 同事们\textcolor{negcolor}{\textbf{不常}}点外卖。 \gloss{(rarely)} \\
\tagbox{BASE} Colleagues \textcolor{negcolor}{\textbf{rarely}} order takeout. \\
\intbox\ Colleagues \textcolor{affcolor}{\textbf{often}} order takeout. \\
\midrule
\tagbox{SRC} 他对坐高铁\textcolor{negcolor}{\textbf{不满意}}。 \gloss{(dissatisfied)} \\
\tagbox{BASE} He is \textcolor{negcolor}{\textbf{dissatisfied}} with taking the high-speed rail. \\
\intbox\ He is \textcolor{affcolor}{\textbf{satisfied}} with the high-speed train. \\
\bottomrule
\end{tabularx}
\caption{\textbf{Examples of activation steering flipping morphological and lexical negations on Qwen3-4B.} \textbf{SRC}: Source sentence; \textbf{BASE}: Un-intervened baseline translation; \textbf{INT}: Model output after steering. Colors indicate \textcolor{affcolor}{\textbf{affirmative states}} and \textcolor{negcolor}{\textbf{negations}}.}
\label{fig:lexical_intervention_examples}
\end{table}

\textbf{Morphological negation translation.} These numbers do not necessarily mean the models are weaker at preserving negation in those translation directions. We manually inspected the  ZH$\xrightarrow{}${EN} baselines of models on the evaluation set, and found 89.8\% of `ineligible' examples are translated as antonyms of the predicates instead of explicit negation markers (\textit{\underline{un}clear} instead of \textit{\underline{not} clear}, \textit{\underline{dis}satisfied} over \textit{\underline{not} satisfied},...) -- which our filter cannot cover. Examples are shown in Table~\ref{fig:lexical_intervention_examples}.
We found that subtracting the negation direction still reverses the original meaning in 68.2\% of these examples.

\begin{table}[h]
\centering
\small
\setlength{\tabcolsep}{5pt}
\begin{tabular}{l ccc}
\toprule
\textbf{Model} & \textbf{Ineligible} & \textbf{Morph./Ant.} & \textbf{Flipped} \\
\midrule
Gemma-3-1B   & 11 & 8  & 7 \\
Gemma-3-4B   & 6  & 6  & 5 \\
Gemma-3-12B  & 5  & 4  & 4 \\
Qwen3-4B     & 12 & 11 & 5 \\
Qwen3.5-9B   & 11 & 11 & 6 \\
Llama-3.1-8B & 4  & 4  & 3 \\
\midrule
\textbf{Total}      & \textbf{49} & \textbf{44} & \textbf{30} \\
\textbf{Percentage} & -- & \textbf{89.8\%} & \textbf{68.2\%} \\
\bottomrule
\end{tabular}
\caption{\textbf{Analysis of ineligible ZH negative-source examples on the Mini-minus-4 evaluation set.}
\textit{Morph./Ant.} Ineligible examples that express negation through negative-prefix morphology or a lexical antonym.
\textit{Flipped} Morph./Ant. cases where intervention successfully flips the baseline output.}
\label{tab:zh_ineligible_analysis}
\end{table}

Nevertheless, we still employ the marker-based eligibility filter for our experiment since these mistakes only affect few ZH-source examples, and it arguably produces more conservative flip rates.

\subsection{Results for Linear Probing}
\subsubsection{Setup}
We use \textbf{linear probing} to test whether negation polarity is linearly encoded in the models' representations. For each model, layer, position, and language, we train a logistic-regression classifier with regularization parameter $C = 1.0$ to predict whether an input sentence is positive or negative.

We further test whether the probe transfers: we evaluate a probe trained on one language's extraction split on the held-out evaluation split of the other languages. 

\subsubsection{Full Probe Accuracy Results}
\label{app:probe}

\begin{figure*}
    \centering
    \includegraphics[width=1\linewidth]{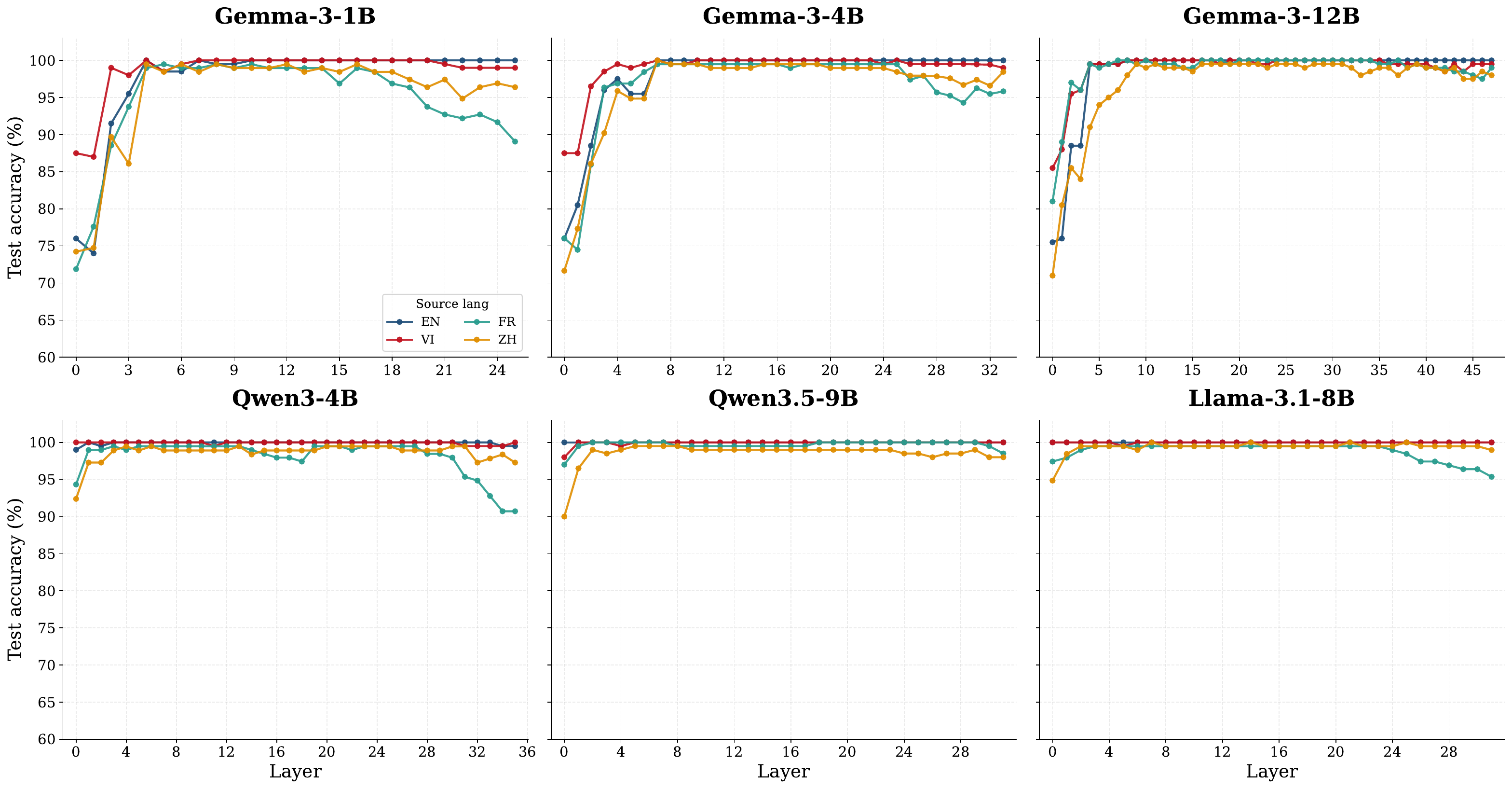}
    \caption{\textbf{Accuracies for probes trained at the NP position.}}
    \label{fig:probe_accuracy_6models_np}
\end{figure*}

\begin{figure*}
    \centering
    \includegraphics[width=1\linewidth]{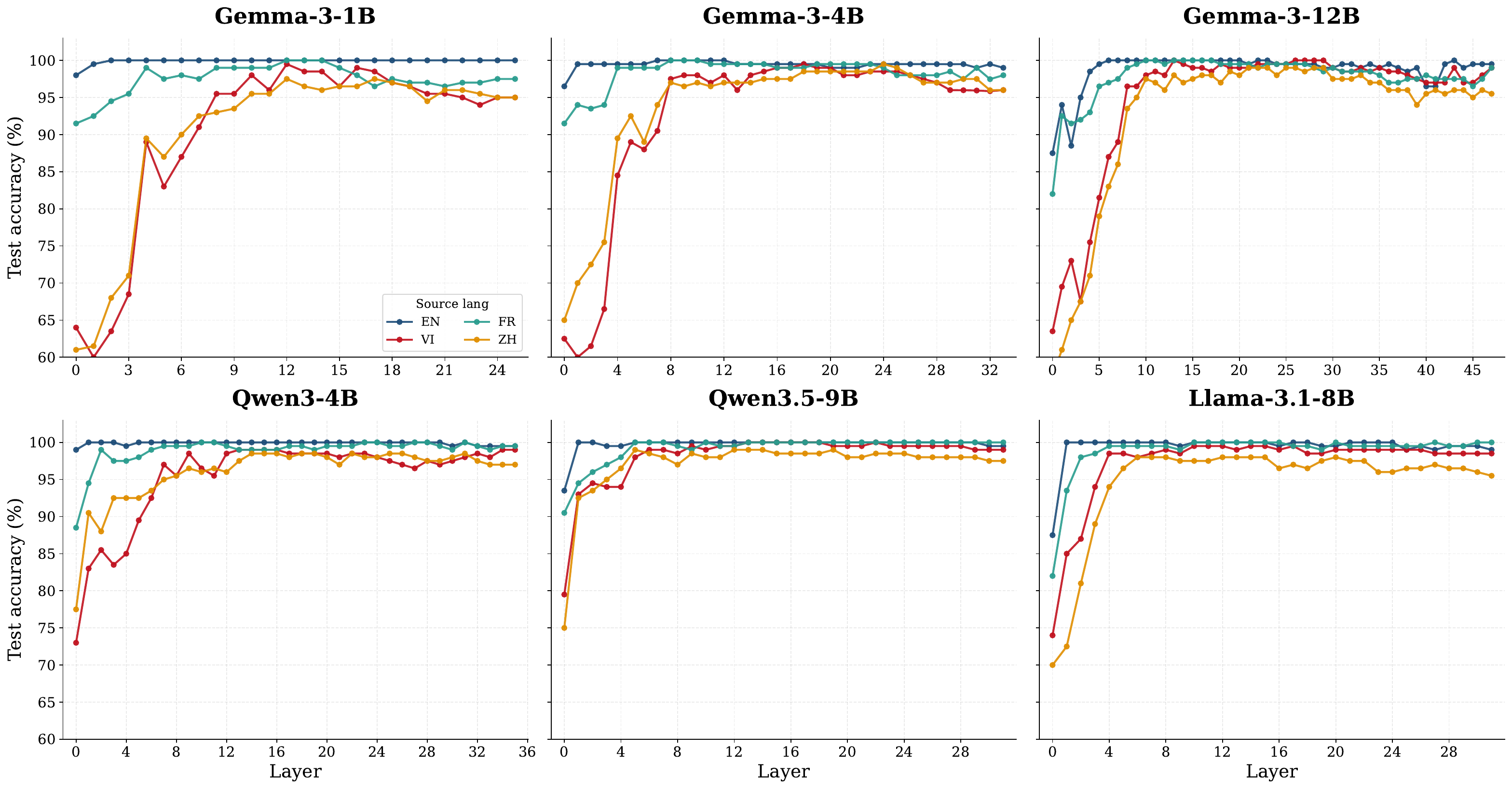}
    \caption{\textbf{Accuracies for probes trained at the FST position.}}
    \label{fig:probe_accuracy_6models_fst}
\end{figure*}

We report the probe accuracy by layer for the NP and FST positions in Figures~\ref{fig:probe_accuracy_6models_np} and \ref{fig:probe_accuracy_6models_fst}. Overall, we find that at both positions, mid-layer probes can predict polarity at perfect accuracy.

\begin{figure*}
    \centering
    \includegraphics[width=1\linewidth]{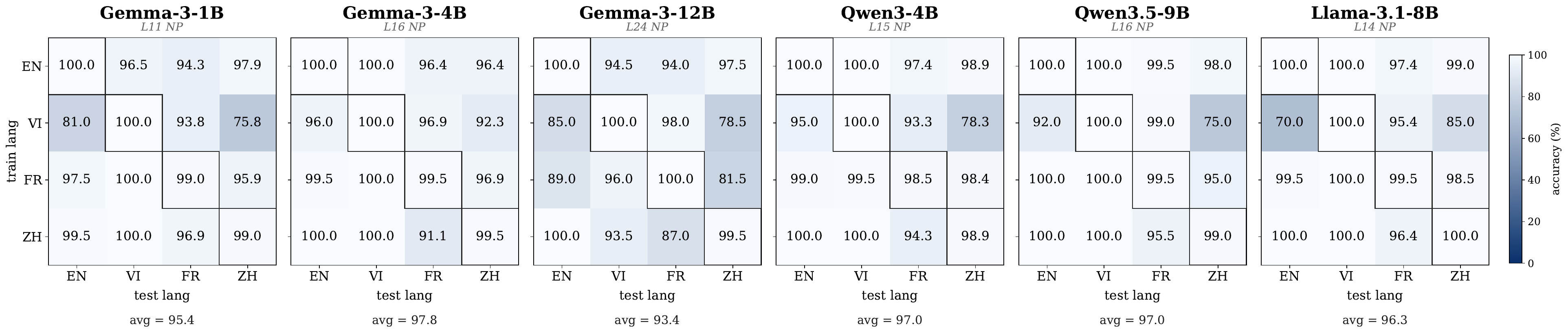}
    \caption{\textbf{Accuracies for cross-language probe transfers.} }
    \label{fig:probe_crosslang_matrices}
\end{figure*}

\subsubsection{Probe Transfer Results}
\label{app:probe_transfer}

We report cross-language probe transfer accuracies in Figure~\ref{fig:probe_crosslang_matrices}. 
For each model, we sweep all layer-position pairs $(l,p)$ and select a single pair based on within-language accuracy. 
At the selected $(l,p)$, we evaluate probes trained in each language on the evaluation sets of all four languages. 
Overall, probes transfer at high accuracy for most language pairs.

\subsection{Extended Results for Contrastive Directions}

\subsubsection{Eval-set Flip Rates for CAA Steering}
\label{app:caa_flip}

\begin{figure*}
    \centering
    \includegraphics[width=1\linewidth]{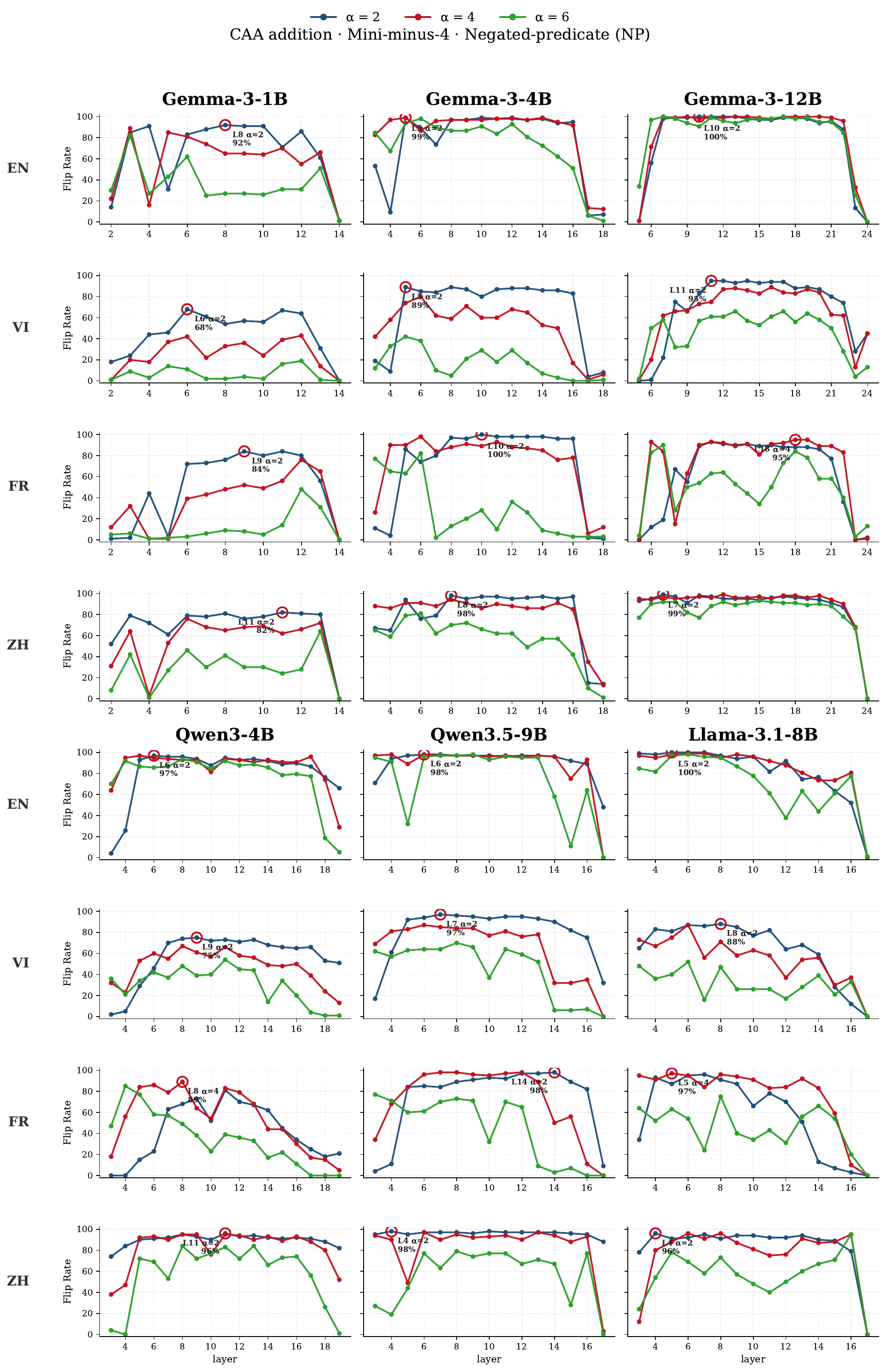}
    \caption{\textbf{Addition rates using the direction extracted at the NP position.}}
    \label{fig:caa_minus4_sweep_np}
\end{figure*}

\begin{figure*}
    \centering
    \includegraphics[width=1\linewidth]{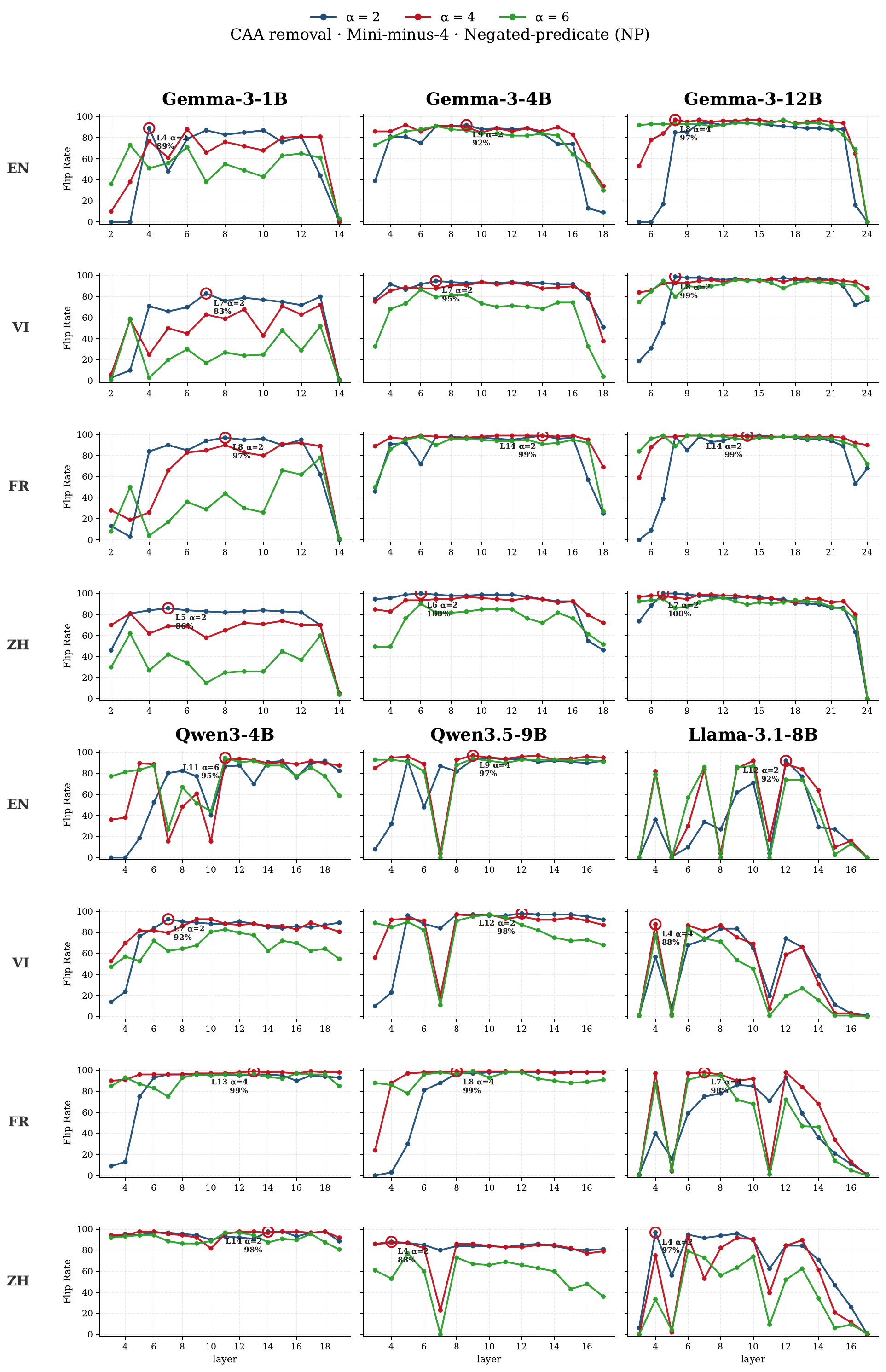}
    \caption{\textbf{Removal rates using the direction extracted at the NP position.}}
    \label{fig:caa_minus4_sweep_np_rem}
\end{figure*}

\begin{figure*}
    \centering
    \includegraphics[width=1\linewidth]{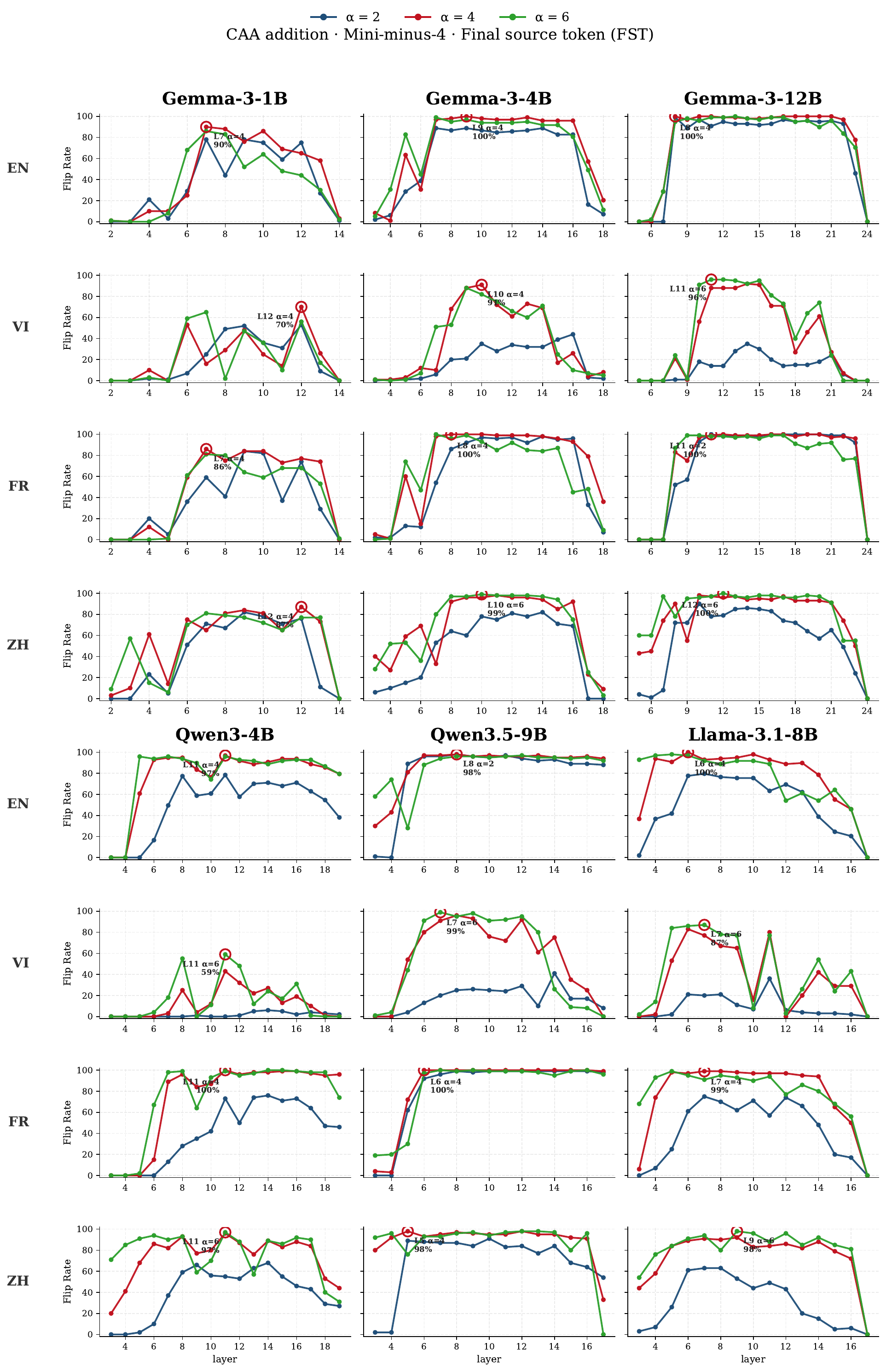}
    \caption{\textbf{Addition rates using the direction extracted at the FST position.}}
    \label{fig:caa_minus4_sweep_fst}
\end{figure*}

\begin{figure*}
    \centering
    \includegraphics[width=1\linewidth]{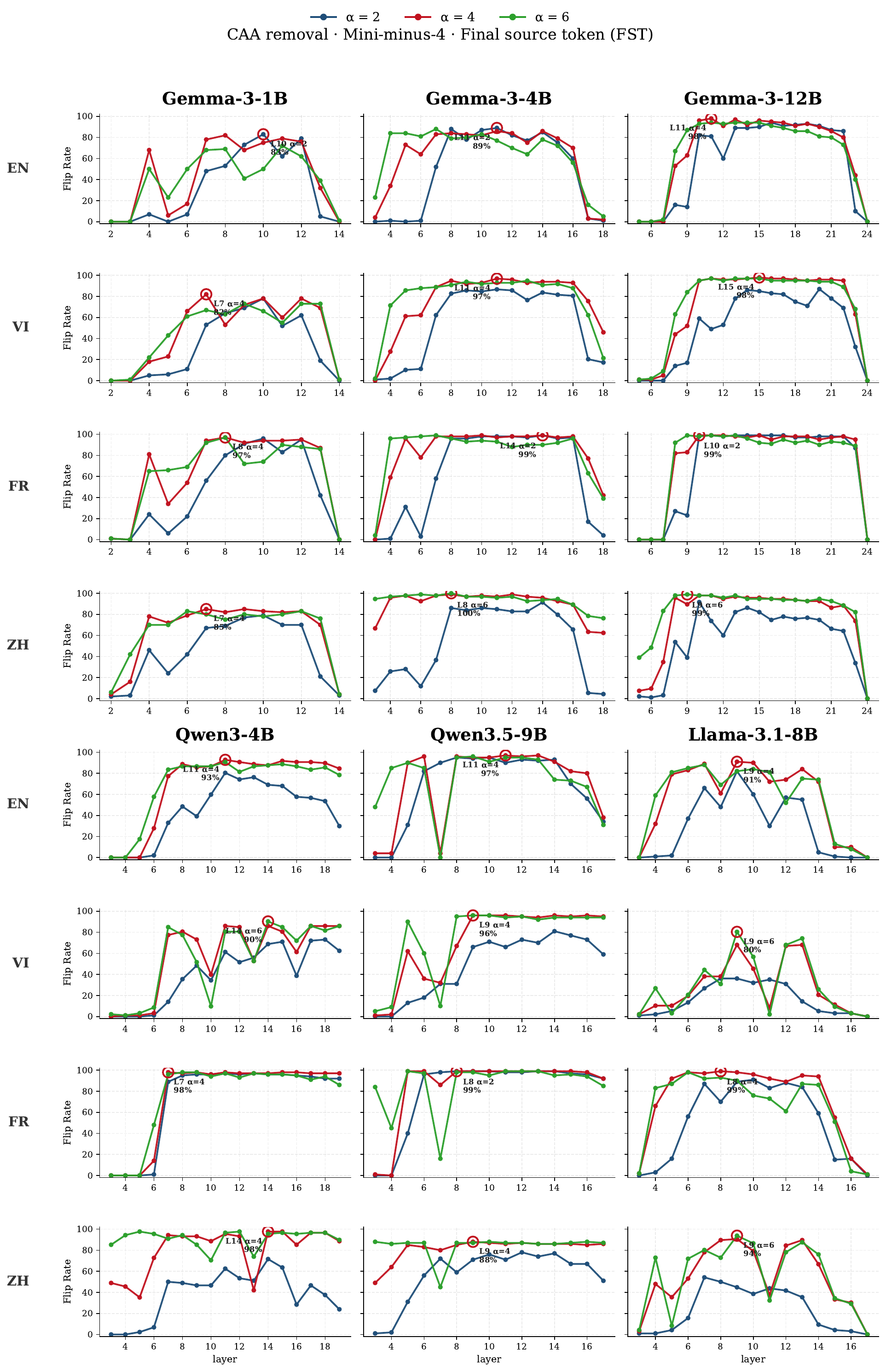}
    \caption{\textbf{Removal rates using the direction extracted at the FST position.}}
    \label{fig:caa_minus4_sweep_fst_rem}
\end{figure*}

We report the eval-set sweep results for every $l$ and $\alpha$ in Figures~\ref{fig:caa_minus4_sweep_np}, \ref{fig:caa_minus4_sweep_np_rem} (NP) and Figures~\ref{fig:caa_minus4_sweep_fst}, \ref{fig:caa_minus4_sweep_fst_rem} (FST). For efficiency, we only sweep 10\% to 55\% of model depth.

We observe a general reverse-U shape, with flip rates rising from early to mid layers and dropping towards middle to later layers.
Notably, while the layer ranges for peak probe accuracy are much wider, the best layers for causal control of negation are much more concentrated, indicating that linear decodability does not equal causal control.
The sweet spot for $\alpha$ is typically 2 or 4, depending on the model and language.

For the within-language experiment in \S\ref{sec:linear}, we choose one representative ($l,p$) based on the best flip rates on the EN set and use the directions at those settings. For each language's direction, we further tune $\alpha$ based on its within-language addition and removal rate on the evaluation set. That is, we select one $\alpha$ for addition and one for removal per language ($\alpha$ values are shown in Figure~\ref{fig:dir_crosslang_transfer}). In case of ties, we select the smaller $\alpha$. These $\alpha$ values are then fixed even when the direction is applied to other languages in the cross-language experiment in \S\ref{sec:crosslingual}.

\begin{table}[h]
\centering
\small
\setlength{\tabcolsep}{4pt}
\begin{tabular}{l ccc}
\toprule
\textbf{Model} & \textbf{$l$} & \textbf{$p$} & \textbf{EN Flip Rate} \\
\midrule
Gemma-3-1B   & L8  & NP & 92.0 / 83.0 \\
Gemma-3-4B   & L10 & NP & 99.0 / 88.0 \\
Gemma-3-12B  & L17 & NP & 100.0 / 97.0 \\
Qwen3-4B     & L13 & NP & 93.8 / 92.8 \\
Qwen3.5-9B   & L9  & NP & 98.0 / 97.0 \\
Llama-3.1-8B & L10 & NP & 95.9 / 92.0 \\
\bottomrule
\end{tabular}
\caption{\textbf{Selected $(l,p)$ configurations and their
EN addition/removal rates.} A single $(l,p)$ is selected
per model for both intervention modes. Rates are reported
as addition/removal.}
\label{tab:model-diag}
\end{table}
                                                                                 
\subsection{Extended Results for SAEs}
\label{app:sae}

\subsubsection{$\alpha^{*}$ used for SAE norm matching}
\label{app:norm-matching}

For SAE norm matching (\S\ref{sec:sae_method}), we use the best EN CAA coefficient at the same layer and position as the selected SAE feature.
At the three SAE layers used in our main experiments, $\alpha_{\mathrm{CAA}}=4$ is optimal on the EN evaluation set (Table~\ref{tab:sae_layer_caa_alpha}).

\begin{table}[h]
\centering
\small
\setlength{\tabcolsep}{3pt}
\begin{tabular}{l c ccc}
\toprule
\textbf{Model} & \textbf{$l$}
& $\boldsymbol{\alpha=2}$
& $\boldsymbol{\alpha=4}$
& $\boldsymbol{\alpha=6}$ \\
\midrule
Gemma-3-4B   & L4  & 9.2 / 81.0  & \textbf{96.9 / 86.0} & 67.3 / 80.0 \\
Qwen3.5-9B   & L10 & 95.0 / 95.0 & \textbf{97.0 / 95.0} & 95.0 / 92.0 \\
Llama-3.1-8B & L13 & 62.2 / 77.0 & \textbf{89.8 / 84.0} & 61.2 / 74.0 \\
\bottomrule
\end{tabular}
\caption{\textbf{EN CAA sweep at the layers of the selected SAE features.}
Each cell reports addition/removal flip rates (\%) on the EN evaluation set.
Bold indicates the $\alpha^{*}$ selected for SAE norm matching.}
\label{tab:sae_layer_caa_alpha}
\end{table}

\subsubsection{Selecting Candidate Layers} 
\label{app:sae_layer_selection}
For each layer of a model, we rank the features by the $\delta_f = \mu_{\mathrm{neg},f} - \mu_{\mathrm{pos},f}$ as described in \S\ref{sec:sae_method}, computed on EN only. Then, we build and steer with the SAE steering vector $\mathbf{d}_\text{SAE}$ using $K$=10 features. Finally, we evaluate flip rates on the EN eval set.

\begin{figure}[t]
    \centering
    \includegraphics[width=1\linewidth]{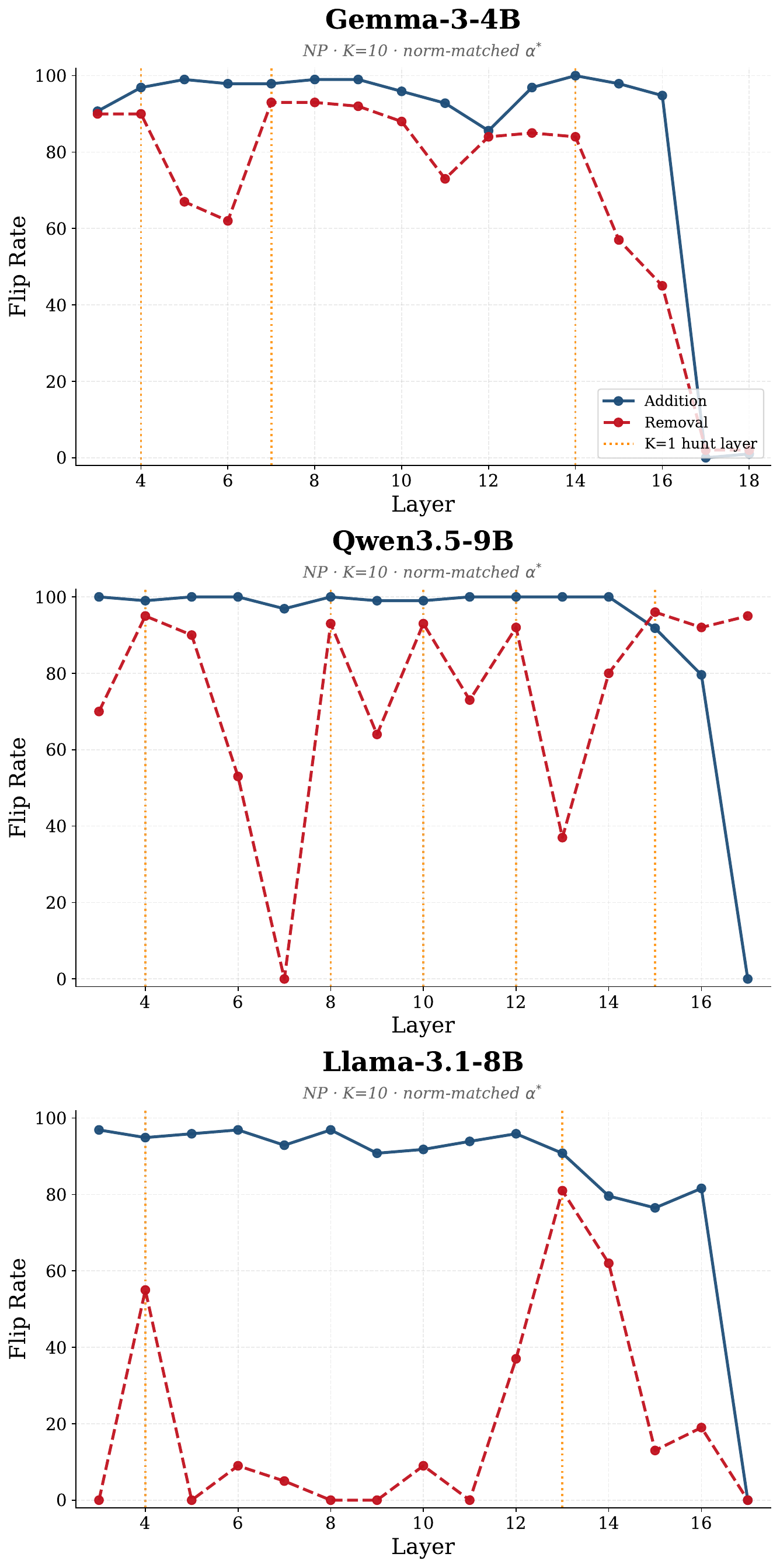}
    \caption{\textbf{Flip rates by layer for $K$=10 SAE features.} Only a few layers per model show the expected effects when steering with SAEs. We choose these layers to search for single effective SAE features.}
    \label{fig:sae_layer_sweep_3models}
\end{figure}

We observe that for each model, only several layers show the expected causal effects when steering with SAE (Figure~\ref{fig:sae_layer_sweep_3models}), unlike CAA steering which is more uniformly successful for EN in the early to middle layers.
Namely, the effective layers for SAE steering are: Gemma-3-4B's $l\in$\{4,7,14\}; Qwen3.5-9B's $l\in$\{4,8,10,12,15\}; Llama-3.1-8B's $l\in$\{4,13\} (marked as dotted lines on the figure). 

\subsubsection{Selecting Features} 
\label{app:sae_feature_selection}
\begin{figure}[t]
    \centering
    \includegraphics[width=1\linewidth]{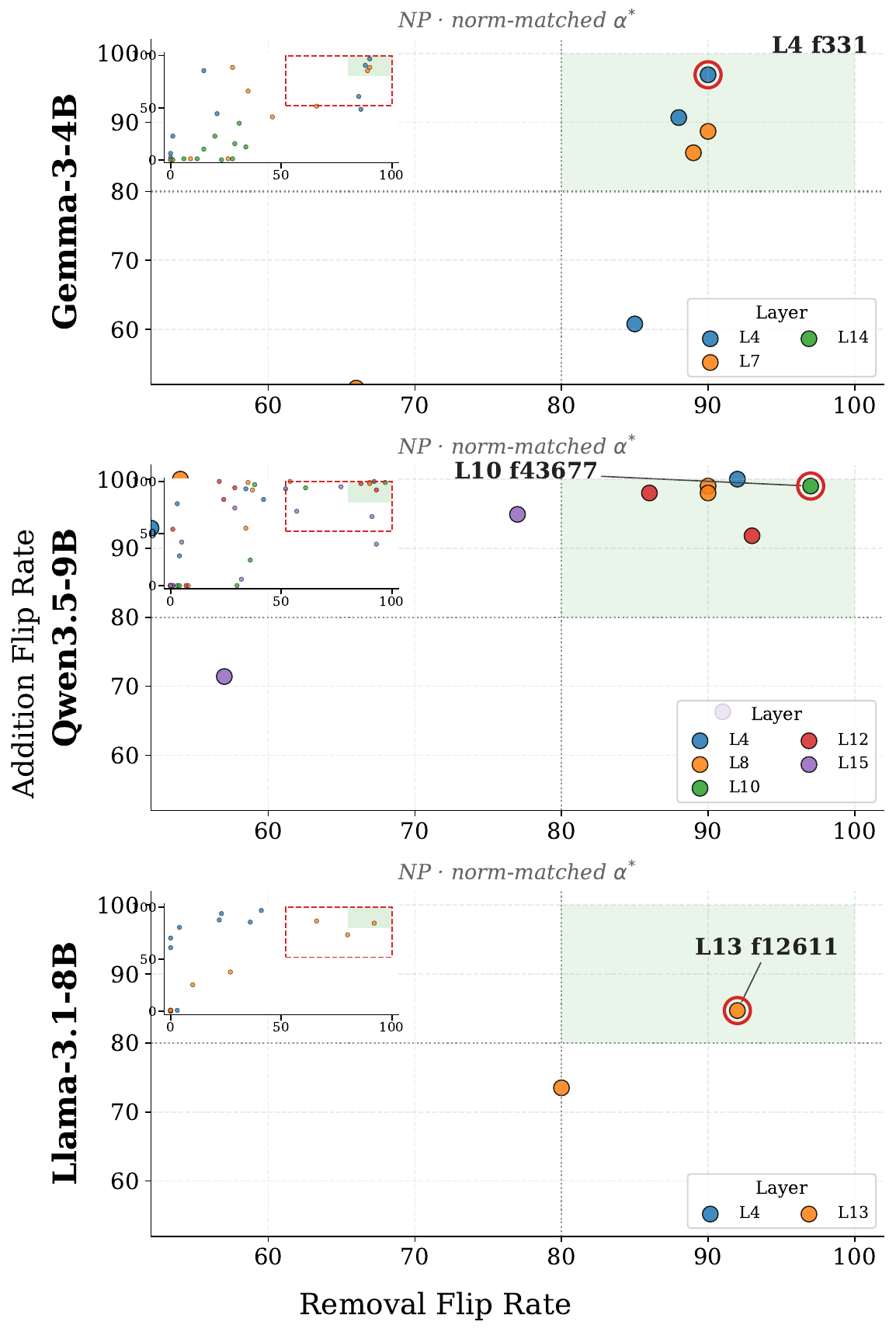}
    \caption{\textbf{Flip rates for individual SAE features at selected layers.} Every model contains at least one feature exceeding 80\% addition/removal rates.
    }
    \label{fig:sae_k1_hunt_bidirectional}
\end{figure}

On these selected layers, we choose the top-10 ranked features per layer and steer with their individual SAE decoder vectors on EN. 
For each model, only a small fraction of tested features exhibit strong polarity control (Figure~\ref{fig:sae_k1_hunt_bidirectional}). 

We choose one feature with the highest addition/removal rates from each model (Gemma's L4 f331, Qwen's L10 f43677, Llama's L13 f12611 – top right of the plot) to evaluate on the test set in \S\ref{sec:sae} and the generalization test in \S\ref{sec:generalize}.

\subsubsection{Feature Interpretation}
\label{app:sae_feature_interpretation}

We further inspect the source-token activations of the three selected $K=1$ features on the Mini-minus-4 extraction set.
For each feature, we collect all non-BOS token positions with positive SAE activation across 150 negative and 150 affirmative examples per language.
Table~\ref{tab:sae_top_tokens} shows representative frequent activating tokens; each count reports the number of firing occurrences in negative / affirmative examples. 

Across all languages, we find that each feature primarily responds to cues related to negation: Gemma-3-4B f331 fires primarily at explicit negation markers and Llama-3.1-8B f12611 predominantly activates on the negated predicates, whereas Qwen3.5-9B f43677 responds to both. 
We further find that Llama's L13 f12611 activates weaker on VI sources compared to other languages, which is one explanation for its weak causal control on this language (\S\ref{sec:sae}). 
The few firings on affirmative examples mostly occur at tokens occupying the negator's position (e.g., Gemma f331 on 很 in ZH) and on \textit{tôi} (VI).

\begin{table*}[t]
    \centering
    \fontsize{8.5pt}{10pt}\selectfont
    \setlength{\tabcolsep}{4pt}
    \renewcommand{\arraystretch}{1.25}

    \begin{tabularx}{\textwidth}{
        @{}
        l
        >{\raggedright\arraybackslash}X
        >{\raggedright\arraybackslash}X
        >{\raggedright\arraybackslash}X
        >{\raggedright\arraybackslash}X
        @{}
    }
        \toprule
        \rowcolor{headergray}
        \textbf{Feature} & \textbf{EN} & \textbf{VI} & \textbf{FR} & \textbf{ZH} \\
        \midrule

        Gemma L4 f331
        & \textbf{not} (132/0); \textbf{cannot} (18/0); \textbf{the} (23/3)
        & \textbf{không} (150/0); \textbf{tôi} (29/29); \textbf{Tôi} (15/15)
        & \textbf{pas} (150/0); \textbf{ne} (118/0); \textbf{n} (32/0)
        & \textbf{不} (78/0); \textbf{我不} (18/0); \textbf{喜欢} (11/0); \textbf{很} (3/41) \\

        \rowcolor{rowgray}
        Qwen L10 f43677
        & \textbf{not} (132/0); \textbf{.} (107/0); \textbf{the} (37/0)
        & \textbf{không} (150/0); \textbf{đọc} (13/0); \textbf{.} (11/0)
        & \textbf{pas} (150/0); \textbf{.} (125/0); \textbf{ne} (118/0)
        & \textbf{不} (37/0); \textbf{。} (18/0); \textbf{不爱} (11/0); \textbf{喜欢} (8/0) \\

        Llama L13 f12611
        & \textbf{.} (83/0); \textbf{the} (19/0); \textbf{read} (5/0); \textbf{easy} (2/0)
        & \textbf{thích} (3/0); \textbf{đủ} (1/0); \textbf{nghe} (1/0)
        & \textbf{pas} (69/0); \textbf{.} (31/0); \textbf{le} (9/0)
        & \textbf{喜欢} (18/0); \textbf{爱} (13/0); \textbf{。} (12/0); \textbf{常} (10/0) \\

        \bottomrule
    \end{tabularx}

    \caption{\textbf{Representative activating tokens for the selected $K=1$ SAE features.}
    Parenthesized counts denote firing occurrences in negative/affirmative examples.
    Gemma-3-4B f331 responds primarily at explicit negation markers;
    Qwen3.5-9B f43677 responds both at negation markers and downstream tokens;
    Llama-3.1-8B f12611 predominantly responds to tokens in negated contexts.}
    \label{tab:sae_top_tokens}
\end{table*}

\section{Relationship between CAA and SAE Directions}
\label{app:caa_sae_relationship}

The SAE direction can be viewed as an approximation of the contrastive activation direction using SAE decoder features. 
Recall that, for a residual activation $\mathbf{h}^{l,p,i}$ with SAE feature
activations $\mathbf{z}^{l,p,i}$, the SAE reconstruction is
\[
\hat{\mathbf{h}}^{l,p,i}
=
\sum_{f=1}^{F}
z_f^{l,p,i}\mathbf{W}_{\mathrm{dec}}[f]
+
\mathbf{b}_{\mathrm{dec}},
\]
where $F=d_{\mathrm{SAE}}$. We use
$\hat{\mathbf{h}}_{\mathrm{neg}}^{l,p,i}$ and
$\hat{\mathbf{h}}_{\mathrm{pos}}^{l,p,i}$ to denote the reconstructions
of the negative and positive members of pair $i$, respectively.

Recall also that
\[
\delta_f
=
\mu_{\mathrm{neg},f}
-
\mu_{\mathrm{pos},f},
\]
where
\[
\mu_{\mathrm{neg},f}
=
\frac{1}{N}\sum_{i=1}^{N}
z_{\mathrm{neg},f}^{l,p,i},
\qquad
\mu_{\mathrm{pos},f}
=
\frac{1}{N}\sum_{i=1}^{N}
z_{\mathrm{pos},f}^{l,p,i}.
\]

If $K=F$, i.e. all features are selected, then
\[
\begin{aligned}
\mathbf{d}_{\mathrm{SAE}}^{l,p,L,F}
&=
\sum_{f=1}^{F}
\delta_f\mathbf{W}_{\mathrm{dec}}[f] \\
&=
\sum_{f=1}^{F}
\left(
\frac{1}{N}\sum_{i=1}^{N}z_{\mathrm{neg},f}^{l,p,i}
-
\frac{1}{N}\sum_{i=1}^{N}z_{\mathrm{pos},f}^{l,p,i}
\right)
\mathbf{W}_{\mathrm{dec}}[f] \\
&=
\frac{1}{N}\sum_{i=1}^{N}
\left(
\hat{\mathbf{h}}_{\mathrm{neg}}^{l,p,i}
-
\hat{\mathbf{h}}_{\mathrm{pos}}^{l,p,i}
\right),
\end{aligned}
\]
where the decoder bias cancels between the negative and positive
reconstructions. 

Thus, $\mathbf{d}_{\mathrm{SAE}}^{l,p,L,F}$ is a reconstruction of the contrastive direction computed from SAE-reconstructed residual activations, 
whereas
\[
\mathbf{d}_{\mathrm{CAA}}^{l,p,L}
=
\frac{1}{N}\sum_{i=1}^{N}
\left(
\mathbf{h}_{\mathrm{neg}}^{l,p,i}
-
\mathbf{h}_{\mathrm{pos}}^{l,p,i}
\right)
\]
is computed from the original residual activations.

Let
\[
\mathbf{e}^{l,p,i}
=
\mathbf{h}^{l,p,i}
-
\hat{\mathbf{h}}^{l,p,i}
\]
denote the SAE reconstruction error. The two directions are therefore
related by
\[
\mathbf{d}_{\mathrm{SAE}}^{l,p,L,F}
=
\mathbf{d}_{\mathrm{CAA}}^{l,p,L}
-
\frac{1}{N}\sum_{i=1}^{N}
\left(
\mathbf{e}_{\mathrm{neg}}^{l,p,i}
-
\mathbf{e}_{\mathrm{pos}}^{l,p,i}
\right).
\]
The difference between the full SAE direction and CAA is
determined by the difference in SAE reconstruction error between the negative and positive examples.

In our experiments, only the features with the largest $\delta_f$ are retained, i.e. $K \ll F$.
The resulting $\mathbf{d}_{\mathrm{SAE}}^{l,p,L,K}$ is therefore a
truncated reconstruction of the same contrast using a small number of
SAE features. 



\end{document}